\documentclass[10pt,twocolumn,letterpaper]{article}
\usepackage{latexml}
\usepackage[pagenumbers]{cvpr}
\usepackage{pifont} %

\newcommand{\projectname}{RIFT}

\newif\ifshowdraftnotes
\showdraftnotesfalse

\definecolor{draftnotecolor}{HTML}{D92B2B}

\usepackage{microtype}

\renewcommand{\paragraph}[1]{\vspace{.5em}\noindent\textbf{#1}}

\definecolor{arrowupcolor}{HTML}{1E8E3E}
\definecolor{arrowdowncolor}{HTML}{D93025}
\newcommand{\uparr}{\textcolor{arrowupcolor}{\boldmath$\uparrow$}}
\newcommand{\dnarr}{\textcolor{arrowdowncolor}{\boldmath$\downarrow$}}
\usepackage{colortbl} %
\usepackage{multirow} %
\usepackage{array}
\usepackage{tabularx} %

\definecolor{cvprblue}{rgb}{0.21,0.49,0.74}
\usepackage[pagebackref,breaklinks,colorlinks,allcolors=cvprblue]{hyperref}
\usepackage{fancyhdr}
\usepackage[most]{tcolorbox}
\usepackage{fontawesome5}
\usepackage{capt-of}

\definecolor{riftgold}{HTML}{C8963E}
\definecolor{riftaccent}{HTML}{9A6A1F}
\definecolor{riftpaper}{HTML}{FBF6EC}

\definecolor{rifttabhi}{HTML}{F2ECE1}    %
\definecolor{rifttabgroup}{HTML}{F6E9CC} %
\definecolor{rifttabalt}{HTML}{FAF7F2}   %
\definecolor{rifttabrule}{HTML}{4A4036}  %
\definecolor{riftyes}{HTML}{2F7D6D}
\definecolor{riftpartial}{HTML}{C8963E}
\definecolor{riftno}{HTML}{B04A3C}
\definecolor{arrowupcolor}{HTML}{2F7D6D}
\definecolor{arrowdowncolor}{HTML}{B04A3C}
\AtBeginDocument{\arrayrulecolor{rifttabrule}}

\newcommand{\riftProjectURL}{https://rift-team.github.io}
\newcommand{\riftCodeURL}{https://github.com/JunhaoWu7/RIFT}
\newcommand{\riftDataURL}{https://huggingface.co/datasets/JunhaoWu7/RIFT}
\newcommand{\riftBoardURL}{https://rift-team.github.io/leaderboard.html}

\makeatletter
\def\rift@nohyph{\raggedright\hyphenpenalty=10000\exhyphenpenalty=10000\relax}
\def\cvprsection{\@startsection{section}{1}{\z@}
   {-10pt plus -2pt minus -2pt}{7pt}{\large\bf\color{riftaccent}\rift@nohyph}}
\def\cvprsubsection{\@startsection{subsection}{2}{\z@}
   {-8pt plus -2pt minus -2pt}{5pt}{\elvbf\color{riftaccent}\rift@nohyph}}

\renewenvironment{abstract}{%
  \begin{tcolorbox}[enhanced, breakable, frame hidden, colback=riftpaper, arc=0pt,
    borderline west={2.2pt}{0pt}{riftgold}, left=10pt, right=10pt, top=5pt, bottom=6pt]
  {\large\bfseries\color{riftaccent}Abstract\par}\vspace{3pt}%
  \small\noindent\ignorespaces}{\end{tcolorbox}}

\newcommand{\riftlink}[4]{%
  \edef\rift@url{#4}%
  \ifx\rift@url\@empty{\color{#1}#2\hspace{0.3em}#3\ifshowdraftnotes\textsuperscript{\color{draftnotecolor}\,TODO}\fi}%
  \else\href{\rift@url}{\color{#1}#2\hspace{0.3em}#3}\fi}

\def\@maketitle{%
  \newpage\null
  \vskip 0.05in
  \begin{center}
    {\Large\bf \@title\par}
    \vspace*{12pt}
    {\large\lineskip .4em
      \begin{tabular}[t]{c}\@author\end{tabular}\par}
    \vspace*{2pt}
  \end{center}}
\makeatother

\makeatletter
\newsavebox{\rift@imgbox}
\newcommand{\riftroundimage}[3][8pt]{%
  \sbox{\rift@imgbox}{\includegraphics[width=#2]{#3}}%
  \begin{tikzpicture}
    \clip[rounded corners=#1] (0,0) rectangle (\wd\rift@imgbox,\ht\rift@imgbox);
    \node[anchor=south west, inner sep=0pt] at (0,0) {\usebox{\rift@imgbox}};
  \end{tikzpicture}}
\makeatother

\newcommand{\riftinstitutions}{%
  \includegraphics[height=21pt]{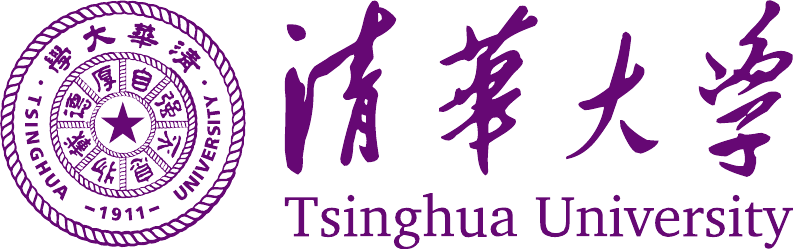}\hspace{16pt}%
  \includegraphics[height=21pt]{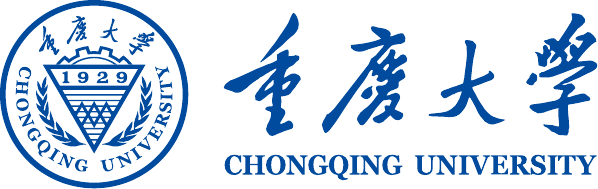}\hspace{16pt}%
  \includegraphics[height=21pt]{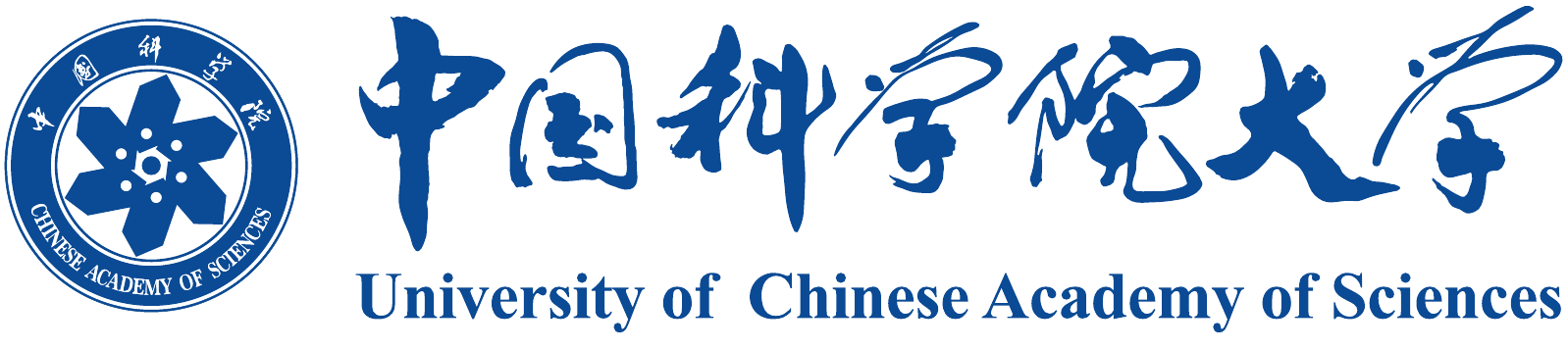}}

\definecolor{riftlinkblue}{HTML}{2B4C7E}
\definecolor{rifthforange}{HTML}{E8920C}
\definecolor{riftgithub}{HTML}{24292F}

\newcommand{\riftlinks}{{\sffamily\bfseries\normalsize
  \riftlink{riftlinkblue}{\faGlobe}{Project Page}{\riftProjectURL}\riftlinksep
  \riftlink{riftgithub}{\faGithub}{GitHub}{\riftCodeURL}\riftlinksep
  \riftlink{rifthforange}{\raisebox{-0.2em}{\includegraphics[height=1.1em]{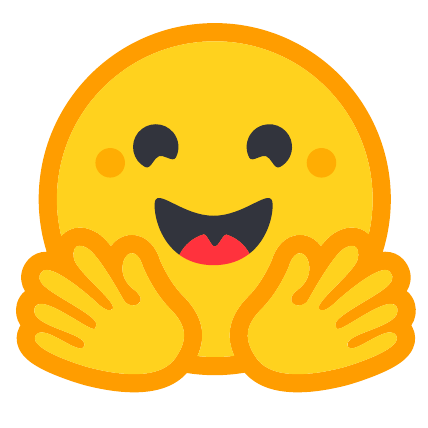}}}{Hugging Face}{\riftDataURL}\riftlinksep
  \riftlink{riftgold}{\faTrophy}{Leaderboard}{\riftBoardURL}}}

\fancypagestyle{riftfirst}{%
  \fancyhf{}%
  \fancyhead[L]{\raisebox{0pt}[15pt][0pt]{\includegraphics[height=21pt]{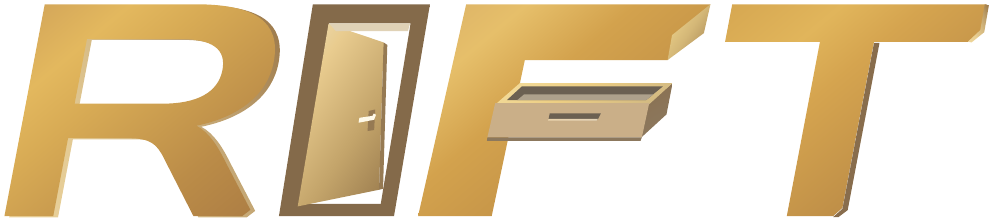}}}%
  \fancyhead[R]{\raisebox{0pt}[15pt][0pt]{\riftinstitutions}}%
  \fancyfoot[C]{\thepage}}

\iflatexml
\renewcommand{\riftlinks}{{\sffamily\bfseries\large
  🌐~\href{\riftProjectURL}{Project Page}\quad$\cdot$\quad
  💻~\href{\riftCodeURL}{GitHub}\quad$\cdot$\quad
  🤗~\href{\riftDataURL}{Hugging Face}\quad$\cdot$\quad
  🏆~\href{\riftBoardURL}{Leaderboard}}}
\newcommand{\riftHTMLfrontmatter}{%
  \begin{center}
  {\large Junhao Wu\textsuperscript{1*}\qquad Haoyun Yang\textsuperscript{1,2*}\qquad Yixiao Huo\textsuperscript{1}\qquad Alex Zhou\textsuperscript{1,3}\qquad Yongchao Chen\textsuperscript{1\textdagger}}\\[0.4em]
  {\textsuperscript{1}Tsinghua University\quad
  \textsuperscript{2}Chongqing University\quad
  \textsuperscript{3}University of Chinese Academy of Sciences\newline
  \textsuperscript{*}Equal~contribution\qquad
  \textsuperscript{\textdagger}Corresponding~author}\\[0.9em]
  \riftlinks
  \end{center}}
\newcommand{\riftColSep}[1]{\setlength{\tabcolsep}{5pt}}
\newcommand{\riftGroupFirst}[2]{\multirow{#1}{*}{#2}}
\newcommand{\riftGroupLast}[2]{}
\newcommand{\riftTwoLine}[2]{#1#2}
\else
\newcommand{\riftColSep}[1]{\setlength{\tabcolsep}{#1}}
\newcommand{\riftGroupFirst}[2]{}
\newcommand{\riftGroupLast}[2]{\multirow{-#1}{*}{#2}}
\newcommand{\riftTwoLine}[2]{\begin{tabular}[c]{@{}l@{}}#1\\#2\end{tabular}}
\fi

\title{\projectname{}: How Faithfully Can Agents Reconstruct Interactive Worlds?}
\iflatexml
\author{}
\else
\author{
{\bfseries Junhao Wu$^{1*}$\quad
Haoyun Yang$^{1,2*}$\quad
Yixiao Huo$^{1}$\quad
Alex Zhou$^{1,3}$\quad
Yongchao Chen$^{1\dagger}$}\\[0.4em]
{\normalsize $^{1}$Tsinghua University\quad
$^{2}$Chongqing University\quad
$^{3}$University of Chinese Academy of Sciences}\\[0.2em]
{\small $^{*}$Equal contribution\quad $^{\dagger}$Corresponding author}
}
\fi

\newcommand{\riftOverviewCaption}{\textbf{Overview of \projectname{}.}
\textbf{Top.} Our containerized pipeline evaluates six frontier coding agents
alongside specialized baselines across three simulators. It scores single-image
3D reconstructions along four fidelity dimensions using hidden 3D layout,
depth, and part references.
\textbf{Bottom.} Representative photographs from \projectname{}100 span all
ten indoor scene categories across \emph{Household30} and \emph{Workplace70}.}

\begin{document}
\iflatexml
\maketitle
\riftHTMLfrontmatter
\newcommand{\riftAbstractText}{%
Reconstructing physically interactive 3D environments from a single casual photograph offers a low-cost path to scalable robot learning, yet requires inferring occluded cavities and part kinematics from static outer surfaces.
Unlike open-loop specialized 3D methods, multimodal coding agents can now author simulation code and refine scenes in closed loop.
Yet how faithfully these agents reconstruct interactive worlds remains unclear, as existing benchmarks focus on static visuals, isolated objects, or motion-revealing videos without testing real-robot usability.
We introduce \projectname{}, 100 single-image tasks across ten indoor categories from 88 real laser-scanned rooms, evaluating six frontier coding agents and four single-image 3D methods in isolated containers across MuJoCo, Isaac Sim, and SAPIEN on appearance, structure, interaction, reproducibility, and simulated Franka manipulation.
The strongest coding agent scores 50.72 to 56.62 out of 100, about five times the strongest single-pass baseline, and controlled ablations attribute this edge to 3D perception and error correction rather than scaffolding.
Yet visually convincing and statically stable scenes still often fail under robot contact due to single-view 3D localization errors, dynamic locking, and handle-contact defects.
On a bimanual Franka platform, policies fine-tuned in reconstructed scenes raise zero-shot contact manipulation mean success from 0.0\% to 16.7\%, and to 30.0\% after multi-dimensional calibration.}
\iflatexml
\section*{Abstract}
\riftAbstractText
\else
\begin{abstract}
\riftAbstractText
\end{abstract}
\fi

\begin{figure*}[t]
  \centering
  \includegraphics[width=\linewidth]{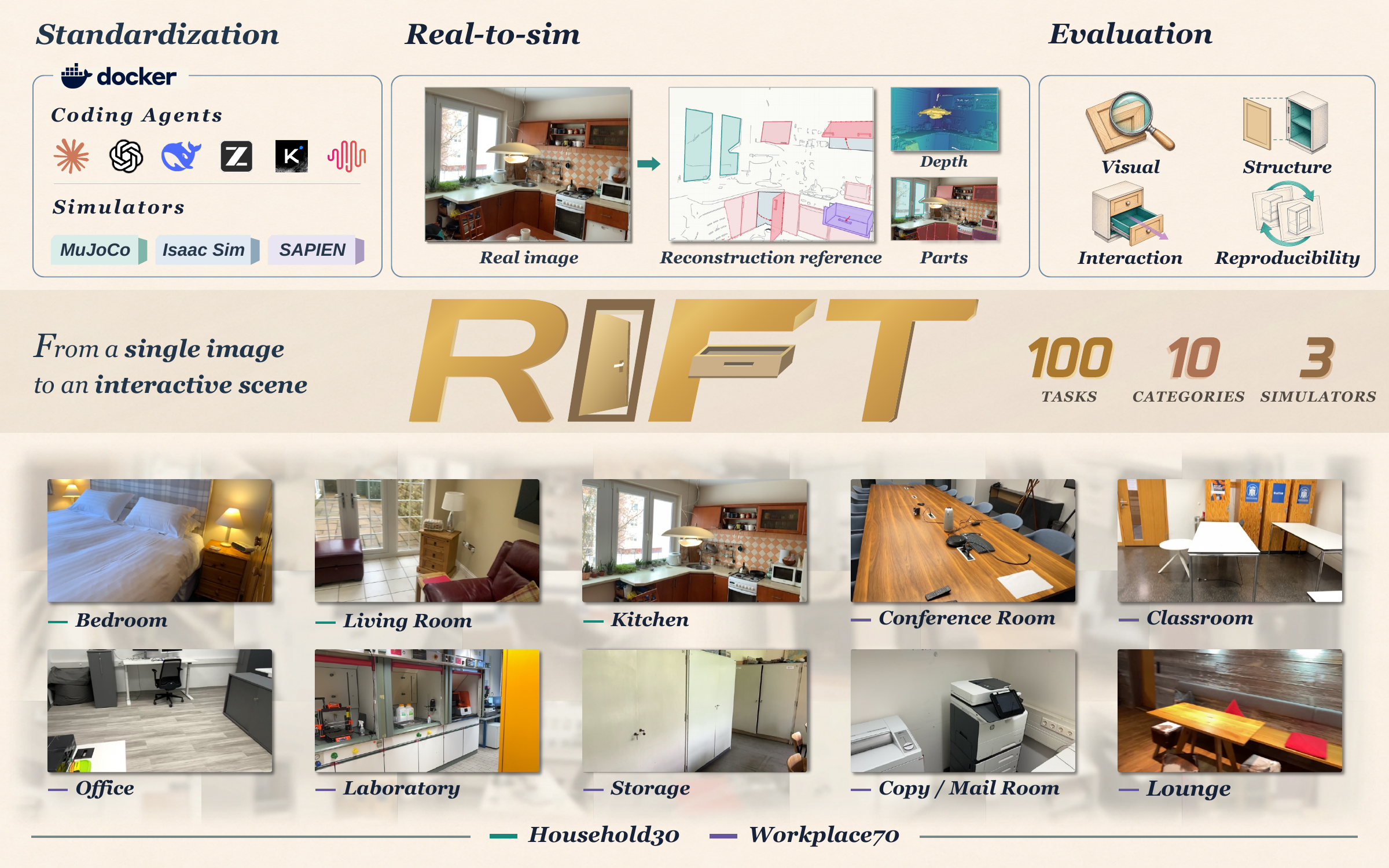}
  \caption{\riftOverviewCaption}
  \label{fig:rift-overview}
\end{figure*}
\else
\twocolumn[{%
  \renewcommand\twocolumn[1][]{#1}%
  \maketitle
  \thispagestyle{riftfirst}
  \begin{center}
    \riftlinks\\[0.7em]
    \begin{minipage}{0.9\textwidth}
    \setlength{\emergencystretch}{1em}%
    \riftroundimage{\linewidth}{figures/RIFT-picture.pdf}
  \captionof{figure}{\riftOverviewCaption}
    \label{fig:rift-overview}
    \end{minipage}
  \end{center}
  \vspace{-1.1em}
  \begin{center}
    \begin{minipage}{0.9\textwidth}
    \newcommand{\riftAbstractText}{%
Reconstructing physically interactive 3D environments from a single casual photograph offers a low-cost path to scalable robot learning, yet requires inferring occluded cavities and part kinematics from static outer surfaces.
Unlike open-loop specialized 3D methods, multimodal coding agents can now author simulation code and refine scenes in closed loop.
Yet how faithfully these agents reconstruct interactive worlds remains unclear, as existing benchmarks focus on static visuals, isolated objects, or motion-revealing videos without testing real-robot usability.
We introduce \projectname{}, 100 single-image tasks across ten indoor categories from 88 real laser-scanned rooms, evaluating six frontier coding agents and four single-image 3D methods in isolated containers across MuJoCo, Isaac Sim, and SAPIEN on appearance, structure, interaction, reproducibility, and simulated Franka manipulation.
The strongest coding agent scores 50.72 to 56.62 out of 100, about five times the strongest single-pass baseline, and controlled ablations attribute this edge to 3D perception and error correction rather than scaffolding.
Yet visually convincing and statically stable scenes still often fail under robot contact due to single-view 3D localization errors, dynamic locking, and handle-contact defects.
On a bimanual Franka platform, policies fine-tuned in reconstructed scenes raise zero-shot contact manipulation mean success from 0.0\% to 16.7\%, and to 30.0\% after multi-dimensional calibration.}
\iflatexml
\section*{Abstract}
\riftAbstractText
\else
\begin{abstract}
\riftAbstractText
\end{abstract}
\fi

    \end{minipage}
  \end{center}
  \vspace{0.8em}
}]
\fi
\section{Introduction}
\label{sec:intro}

Building large-scale, physically interactive 3D simulation environments is foundational for robot policy training, scalable evaluation, and sim-to-real transfer.
Compared with multi-view scanning, depth calibration, or interaction videos, a single casual indoor RGB photograph is ubiquitous and low-cost, offering a scalable entry point for environment generation and interaction data collection.
Yet reconstructing an interactive world from a single static image remains highly challenging, as the photograph typically shows only outer surfaces at rest such as closed cabinet doors and drawers without revealing occluded interior cavities or part motion.
Supporting physical contact manipulation therefore requires moving beyond a visually similar static 3D model to simultaneously recover accurate spatial layout, plausible part kinematics, and reliable contact properties.

Most specialized 3D reconstruction methods follow a single-pass feed-forward paradigm that outputs closed static meshes or assembles preset templates, lacking simulation feedback to resolve interpenetration and jammed parts~\cite{chen2025sam3d,huang2025midi,chen2024urdformer,li2026particulate}.
Meanwhile, general-purpose coding agents with strong visual understanding, spatial reasoning, and long-horizon coding capabilities have advanced rapidly, enabling them to directly author simulation code and refine scenes in closed loop from runtime feedback.
Such on-demand simulation construction aligns naturally with single-view vision-language-action models and agentic control.
However, how faithfully general-purpose coding agents can reconstruct interactive worlds, and how much advantage they hold over traditional 3D reconstruction methods, still lacks a reliable benchmark.
Most existing benchmarks stop at static visual evaluation without physical checks~\cite{li2026legoanything,lin2026frontier3dbench}, while the few that test physics remain confined to isolated objects or motion-revealing videos~\cite{cai2026artisan,shao2026physcodegym,tang2026video2world}, falling short of fairly comparing both paradigms and verifying downstream robot manipulation and real-world transfer.

\begin{table*}[t]
  \centering
  \caption{\textbf{Comparison of \projectname{} with representative interactive reconstruction methods and scene-coding benchmarks.}
  Rows are grouped into specialized reconstruction pipelines and coding-agent benchmarks, where \textcolor{riftyes}{\ding{51}} indicates explicit coverage, \textcolor{riftpartial}{$\bullet$} partial coverage, and \textcolor{riftno}{\ding{55}} not reported.
  Full coverage under the fidelity and diagnosis columns requires direct quantitative evaluation against ground truth, contact-physics verification beyond kinematics, and categorized failure breakdowns, respectively.
  \emph{Cross-Simulator} and \emph{Cross-Paradigm} denote evaluation across multiple physics engines and unified benchmarking of specialized methods alongside coding agents, while \emph{Real-World Utility} indicates validation on physical robot hardware.}
  \label{tab:related-work}
  \footnotesize
  \riftColSep{1pt}
  \renewcommand{\arraystretch}{1.12}
  \begin{tabularx}{\textwidth}{@{}ll*{7}{>{\centering\arraybackslash}X}@{}}
    \toprule
    \textbf{Work} & \textbf{Input}
    & \textbf{\shortstack{Visual\\Fidelity}}
    & \textbf{\shortstack{Structural\\Fidelity}}
    & \textbf{\shortstack{Interaction\\Fidelity}}
    & \textbf{\shortstack{Error\\Diagnosis}}
    & \textbf{\shortstack{Cross-\\Simulator}}
    & \textbf{\shortstack{Cross-\\Paradigm}}
    & \textbf{\shortstack{Real-World\\Utility}} \\
    \midrule
    URDFormer~\cite{chen2024urdformer}
    & Single image
    & \textcolor{riftno}{\ding{55}} & \textcolor{riftyes}{\ding{51}} & \textcolor{riftpartial}{$\bullet$} & \textcolor{riftno}{\ding{55}} & \textcolor{riftno}{\ding{55}} & \textcolor{riftno}{\ding{55}} & \textcolor{riftyes}{\ding{51}} \\
    DRAWER~\cite{xia2025drawer}
    & Video
    & \textcolor{riftyes}{\ding{51}} & \textcolor{riftyes}{\ding{51}} & \textcolor{riftpartial}{$\bullet$} & \textcolor{riftpartial}{$\bullet$} & \textcolor{riftno}{\ding{55}} & \textcolor{riftno}{\ding{55}} & \textcolor{riftyes}{\ding{51}} \\
    \midrule
    BVB~\cite{tang2026bvb}
    & Video
    & \textcolor{riftyes}{\ding{51}} & \textcolor{riftpartial}{$\bullet$} & \textcolor{riftno}{\ding{55}} & \textcolor{riftpartial}{$\bullet$} & \textcolor{riftno}{\ding{55}} & \textcolor{riftno}{\ding{55}} & \textcolor{riftno}{\ding{55}} \\
    SceneActBench~\cite{zhao2026sceneactbench}
    & Multi-view + cameras
    & \textcolor{riftyes}{\ding{51}} & \textcolor{riftyes}{\ding{51}} & \textcolor{riftpartial}{$\bullet$} & \textcolor{riftpartial}{$\bullet$} & \textcolor{riftno}{\ding{55}} & \textcolor{riftno}{\ding{55}} & \textcolor{riftno}{\ding{55}} \\
    LEGO-Bench~\cite{li2026legoanything}
    & Single image + cameras
    & \textcolor{riftyes}{\ding{51}} & \textcolor{riftyes}{\ding{51}} & \textcolor{riftpartial}{$\bullet$} & \textcolor{riftyes}{\ding{51}} & \textcolor{riftno}{\ding{55}} & \textcolor{riftno}{\ding{55}} & \textcolor{riftno}{\ding{55}} \\
    PhysCodeGym~\cite{shao2026physcodegym}
    & Video
    & \textcolor{riftyes}{\ding{51}} & \textcolor{riftpartial}{$\bullet$} & \textcolor{riftyes}{\ding{51}} & \textcolor{riftyes}{\ding{51}} & \textcolor{riftno}{\ding{55}} & \textcolor{riftno}{\ding{55}} & \textcolor{riftno}{\ding{55}} \\
    Video2World~\cite{tang2026video2world}
    & Embodied video
    & \textcolor{riftpartial}{$\bullet$} & \textcolor{riftyes}{\ding{51}} & \textcolor{riftyes}{\ding{51}} & \textcolor{riftyes}{\ding{51}} & \textcolor{riftyes}{\ding{51}} & \textcolor{riftno}{\ding{55}} & \textcolor{riftno}{\ding{55}} \\
    \midrule
    \rowcolor{rifttabhi}
    \textbf{\projectname{}, ours}
    & \textbf{Single image}
    & \textcolor{riftyes}{\ding{51}} & \textcolor{riftyes}{\ding{51}} & \textcolor{riftyes}{\ding{51}} & \textcolor{riftyes}{\ding{51}} & \textcolor{riftyes}{\ding{51}} & \textcolor{riftyes}{\ding{51}} & \textcolor{riftyes}{\ding{51}} \\
    \bottomrule
  \end{tabularx}
\end{table*}

To address this gap, we introduce \textbf{\projectname{}}, the Reconstruction and Interaction Fidelity Test for single-image interactive indoor scene reconstruction.
To objectively evaluate hidden spatial structure and kinematic parameters from a single static image, \projectname{} constructs \textbf{100 tasks} across ten household and workplace categories from 88 real laser-scanned indoor rooms, annotating 908 oriented 3D furniture bounding boxes and 703 articulated parts as unified cross-simulator ground truth.
All evaluations run in standardized containers isolated from ground truth under uniform resource quotas across MuJoCo~\cite{todorov2012mujoco}, Isaac Sim~\cite{nvidia2025isaacsim}, and SAPIEN~\cite{xiang2020sapien}, fairly comparing closed-loop coding agents and single-pass baselines.
\projectname{} separately quantifies visual appearance, spatial layout, physical interaction, and code reproducibility, pairing fine-grained failure diagnostics with simulated robot contact manipulation tests to verify operability while offering practical insights toward real-world policy transfer.

In summary, our main contributions are threefold:
\begin{itemize}
  \item \textbf{Cross-Simulator and Cross-Paradigm Benchmark.} We establish \projectname{} to uniformly evaluate general-purpose coding agents and specialized 3D baselines from a single uncalibrated image inside strictly isolated containers across MuJoCo, Isaac Sim, and SAPIEN on visual, structural, interactive, and clean-rebuild fidelity, with automated rankings matching blind pairwise human judgments of scene quality.
  \item \textbf{Paradigm-Gap Discovery and Bottleneck Diagnosis.} Benchmarking six frontier coding agents and four specialized baselines with controlled harness and reasoning-effort ablations, including a Bash-only minimal harness, we show that the strongest closed-loop agent outperforms the best single-pass method by \textbf{about fivefold} via intrinsic 3D perception and error correction rather than outer scaffolding, while top scores \textbf{barely above 50} expose universal bottlenecks in metric 3D grounding and dynamic collider self-locking.
  \item \textbf{Embodied Operability and Zero-Shot Real-World Transfer.} We pair simulated Franka Panda contact tests that isolate reachability and handle slippage with physical bimanual Franka Research 3 trials, raising zero-shot mean success from \textbf{0.0\%} for the unfinetuned policy to \textbf{16.7\%} on raw agent reconstructions and \textbf{30.0\%} after multi-dimensional calibration.
\end{itemize}
\section{Related Work}
\label{sec:related}

\subsection{Single-Image 3D and Interactive Reconstruction}
\label{sec:related-realtosim}
In reconstructing real-world scenes for embodied training and interaction data collection, DRAWER~\cite{xia2025drawer} and PhysTwin~\cite{jiang2025phystwin} rely on multi-view scans or interaction videos to build interactive environments, yet heavy capture workflows limit scalable deployment and motivate single-image inputs.
For single-image reconstruction, URDFormer~\cite{chen2024urdformer} and ACDC~\cite{dai2024acdc} assemble movable scenes by matching parametric articulated templates or retrieving external 3D assets, yet preset models struggle to faithfully recover diverse indoor geometry.
To move beyond preset templates, recent feed-forward methods directly recover 3D structure from images, as SAM 3D Objects~\cite{chen2025sam3d} and MIDI~\cite{huang2025midi} generate object meshes and multi-object layouts while PhysGen3D~\cite{chen2025physgen3d}, SINGAPO~\cite{liu2025singapo}, and Particulate~\cite{li2026particulate} further add collision estimation and part-level joint prediction.
However, lacking runtime feedback from a physics simulator, such single-pass open-loop methods cannot complete occluded cavities and handles or iteratively repair interpenetration, instability, and jammed joints, falling short of building faithful and directly operable simulation environments from a single image.
\vspace{-2pt}
\subsection{Programmatic and Agentic Simulation Construction}
\label{sec:related-code}
To overcome the physical bottlenecks of open-loop reconstruction, recent works represent scene geometry, joints, and dynamics directly as executable code, leveraging foundation models and agents to build simulation environments.
For synthetic scenes, Infinigen Indoors~\cite{raistrick2024infinigenindoors} and Holodeck~\cite{yang2024holodeck} arrange assets via procedural rules or language instructions, while RoboGen~\cite{wang2024robogen} and GenSim2~\cite{hua2024gensim2} prompt multimodal models to author articulated task code, yet all target open-ended virtual generation rather than faithfully reconstructing a specific real photograph.
For observation-grounded reconstruction, Real2Code~\cite{mandi2024real2code}, Articulate-Anything~\cite{le2025articulateanything}, and Articraft~\cite{zhou2026articraft} pair visual feedback with code generation to recover joints, though only for isolated objects.
Although Agentic Real2Sim~\cite{chen2026agenticreal2sim} and SceneAgent~\cite{hollis2026sceneagent} chain segmentation, retrieval, and assembly to reach full scenes, they still rely on multi-view inputs, external 3D asset banks, or custom interfaces.
By contrast, general-purpose multimodal coding agents can bypass external asset libraries and custom pipelines to write native simulation code directly from a single image and debug in closed loop, though their true capability boundaries without explicit 3D shape priors remain unclear.

\subsection{Related Benchmarks}
\label{sec:related-benchmarks}
Systematically evaluating the geometric fidelity and physical interactability of generated scenes has become essential, and Table~\ref{tab:related-work} compares representative methods and benchmarks.
Most existing benchmarks evaluate visual or static geometric quality inside graphics software, spanning open-ended or single-image modeling in Frontier3DBench~\cite{lin2026frontier3dbench} and LEGO-Bench~\cite{li2026legoanything}, posed multi-view scenes in SceneActBench~\cite{zhao2026sceneactbench}, and videos in BVB~\cite{tang2026bvb} and 4DCodeBench~\cite{shen2026fourdecodebench}.
Checking only surface shapes, renders, or scripted replay without physics-simulator verification leaves unstable centers of mass, jammed doors, and missing cavities undetected.
Among physics-based benchmarks, Artisan~\cite{cai2026artisan} evaluates joint motion and manipulation only on isolated objects detached from room layout and clearance constraints, while PhysCodeGym~\cite{shao2026physcodegym} and Video2World~\cite{tang2026video2world} cover full scenes from videos that already reveal motion trajectories rather than inferring occluded interiors and kinematics from a single closed-door photograph.

Existing benchmarks also typically support only a single backend, lack a unified comparison between coding agents and traditional 3D methods, and omit real-robot verification.
As shown in Table~\ref{tab:related-work}, \projectname{} evaluates both classes of methods from a single uncalibrated image across three physics engines, jointly measuring visual, structural, and physical interaction fidelity while localizing fine-grained errors.
Combining simulated manipulation with physical robot trials, it further tests real-world sim-to-real transferability.

\section{\projectname{}: Reconstruction and Interaction Fidelity Test}
\label{sec:benchmark}

\begin{figure}[b]
  \centering
  \includegraphics[width=\columnwidth]{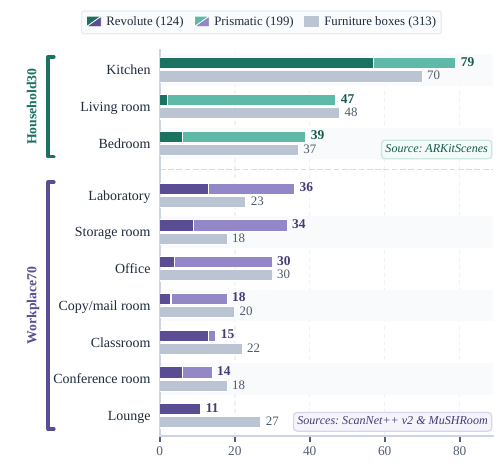}
  \caption{\textbf{Category composition and scored annotation density of \projectname{}100.}
    Ten indoor categories of 10 tasks each are grouped into \emph{Household30} and \emph{Workplace70}.
    Stacked upper bars report 124 revolute and 199 prismatic scored parts, 323 in total, while gray lower bars report 313 scored 3D furniture boxes in total; full statistics appear in Supplementary Sec.~\ref{sec:supp-task-specification}.}
  \label{fig:task-composition}
\end{figure}

\begin{figure*}[t]
  \centering
  \includegraphics[pagebox=cropbox,width=\textwidth]{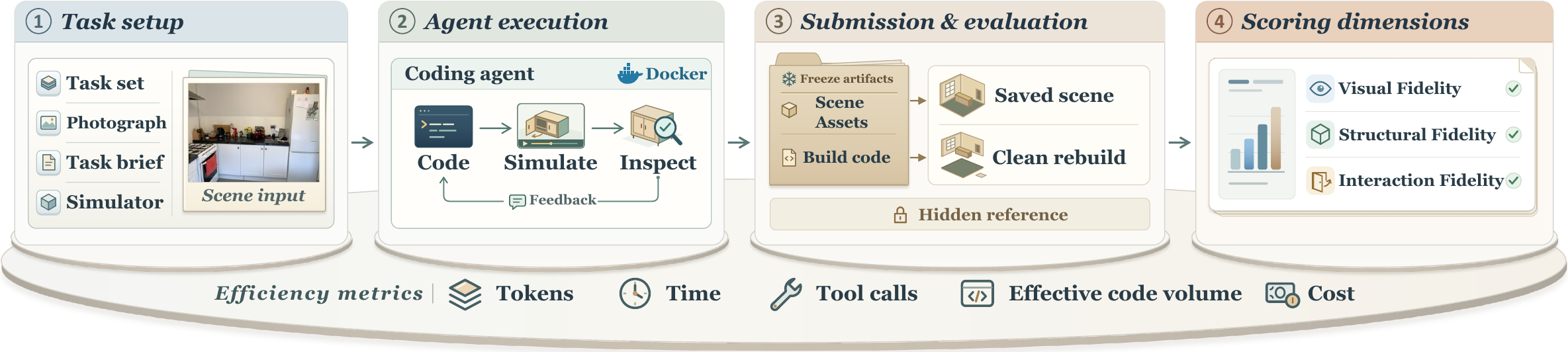}
  \caption{\textbf{Evaluation protocol for coding agents in \projectname{}.}
    \textbf{1.}~Task setup provides the input image, brief, and target simulator. \textbf{2.}~The coding agent iterates via coding, simulation, and inspection inside an isolated Docker container. \textbf{3.}~Both the frozen saved scene and an offline clean rebuild are evaluated against hidden references. \textbf{4.}~Scoring quantifies visual, structural, and interaction fidelity while logging five efficiency metrics, shown at the bottom.}
  \label{fig:rift-evaluation}
\end{figure*}

As illustrated in Figure~\ref{fig:rift-overview}, \projectname{} is a single-image interactive indoor scene reconstruction benchmark spanning MuJoCo~\cite{todorov2012mujoco}, Isaac Sim~\cite{nvidia2025isaacsim}, and SAPIEN~\cite{xiang2020sapien}.
Combining standardized evaluation with simulated robot manipulation tests, it aims to systematically measure general-purpose coding agents and specialized 3D reconstruction methods across visual, structural, physical interaction, and reproducibility dimensions, offering practical insights into a viable path from single-image reconstruction to real-world policy transfer.

\subsection{Task Definition}
\label{sec:task}
Given a single uncalibrated indoor RGB photograph and a target simulator, an evaluated system must reconstruct a visually consistent, physically stable, and interactive 3D scene without depth maps, camera parameters, or external 3D asset libraries.
From the single view, the system must recover camera parameters, metric scale, and 3D layout, complete furniture cavities for gravitational stability, and build accurate joints, colliders, and handles on doors and drawers for interpenetration-free grasping and motion.
Each submission includes a directly loadable scene, a structured semantic specification, and offline-reproducible build code.

\subsection{Task Sets and Reference Construction}
\label{sec:data}

\paragraph{Task sources and composition.}
As shown in Figure~\ref{fig:task-composition}, \projectname{}100 selects 100 single-image tasks from 88 real laser-scanned indoor rooms evenly spanning ten space categories, comprising \emph{Household30} from three residential categories in ARKitScenes~\cite{baruch2021arkitscenes} and \emph{Workplace70} from seven office and institutional categories in ScanNet++ v2~\cite{yeshwanth2023scannetpp} and MuSHRoom~\cite{ren2024mushroom}.
Candidate images are filtered for sharpness, exposure, and camera-to-scan alignment, while laser scans, calibrated camera parameters, and ground-truth annotations are hidden from evaluated systems and used only by the backend evaluator.

\paragraph{Hidden reference annotations.}
Each task pairs with a frozen three-layer reference package shared across all three simulators.
The \textbf{scene geometry layer} projects the laser-scanned mesh through the calibrated camera to produce aligned depth and room-surface references.
The \textbf{object layout layer} refits oriented 3D furniture and appliance boxes against scanned surfaces to centimeter accuracy and annotates tabletop small-item support, totaling 908 3D boxes and 919 support relations with 313 primary visible boxes scored.
The \textbf{part articulation layer} specifies kinematic parameters and handle geometry, refined from SceneFun3D~\cite{delitzas2024scenefun3d} for household scenes and annotated on laser scans with Articulate3D~\cite{halacheva2025articulate3d} cross-checks for workplace scenes, totaling 282 revolute and 421 prismatic joints across 703 movable parts with up to 46 per image and 323 unobstructed camera-facing parts scored.

\paragraph{Quality control.}
All references pass automated geometric checks and multi-stage independent blind reviews before checksum locking, and are validated via progressive benchmark scenes from bare room shells to full articulated ground truth to verify monotonic scorer response, see Supplementary Secs.~\ref{sec:supp-task-specification} and~\ref{sec:supp-references}.

\subsection{Evaluation Protocol and Metrics}
\label{sec:protocol}

\paragraph{Isolated execution and clean rebuild.}
As shown in Figure~\ref{fig:rift-evaluation}, every run executes inside an isolated container under uniform hardware quotas, mounting no ground truth and restricting outbound traffic solely to the designated model inference endpoint.
Agents invoke native tools to author scene code and refine it in closed loop via in-container view rendering, physical stability checks, and public robot self-check scripts.
Upon completion or timeout, the system freezes the saved scene and logs, reruns the submitted script from scratch inside a fresh offline container, and evaluates both artifacts, with resource budgets in Sec.~\ref{sec:setup}.

\paragraph{Composite quality score.}
Table~\ref{tab:metric-definitions} summarizes the seven scoring components corresponding to the three fidelity dimensions in Figure~\ref{fig:rift-evaluation}.
\begin{table}[b]
  \centering
  \caption{\textbf{Scoring components in \projectname{}.}
  $A$, $O$, $L$, and $J$ form the weighted quality term, while physical validity $P$, completeness $C$, and scale correction $\kappa(e)$ act as multiplicative adjustment factors.}
  \label{tab:metric-definitions}
  \small
  \riftColSep{3pt}
  \renewcommand{\arraystretch}{1.1}
  \begin{tabular*}{\columnwidth}{@{\extracolsep{\fill}}ll@{}}
    \toprule
    \textbf{Component} & \textbf{Evaluation target} \\
    \midrule
    Appearance $A$ & Color and edge agreement \\
    Object geometry $O$ & Matching, position, size, shape \\
    Spatial layout $L$ & Depth and floor occupancy \\
    Articulation $J$ & Joint accuracy and motion \\
    \midrule
    Physical validity $P$ & Stability, contacts, densities \\
    Completeness $C$ & Interior and handle geometry \\
    Scale correction $\kappa(e)$ & Global scale error \\
    \bottomrule
  \end{tabular*}
\end{table}

The composite quality score is
\begin{equation}
S = 100\,Q_{\mathbf w}\,P\,\frac{1+C}{2}\,\kappa(e), \qquad Q_{\mathbf w} = \frac{\sum_{i\in\mathcal I} w_i q_i}{\sum_{i\in\mathcal I} w_i},
\label{eq:quality-score}
\end{equation}
where $\mathbf q=(J,O,L,A)$ with weights $\mathbf w=(6,4,4,3)$ prioritizes articulated interaction and 3D geometry over surface appearance, and $\mathcal I$ indexes applicable task terms while omitting $J$ when no scored articulated parts exist.
Multiplicative terms $P$, $(1+C)/2\in[0.5,1]$, and $\kappa(e)\in[0.7,1]$ from the global scale log error $e$ prevent unstable, cavity-missing, or mis-scaled scenes from scoring high via local geometry alone.

\paragraph{Reproducibility, failure diagnostics, and robot manipulation.}
Evaluating the frozen and offline-rebuilt scenes under Eq.~\eqref{eq:quality-score} yields $S_{\text{frozen}}$ and $S_{\text{rebuilt}}$, assigning zero upon load or rebuild failure so their gap measures code reproducibility.
The evaluator also computes eight fine-grained failure rates and runs independent simulated Franka Panda handle-grasping and opening tests across all 323 scored parts to verify contact operability, while logging active time, tokens, tool calls, and code lines, see Supplementary Secs.~\ref{sec:supp-quality} and~\ref{sec:supp-comparisons}.

\begin{table*}[t]
  \centering
  \caption{\textbf{Main reconstruction results across physics simulators and task sets.}
  Each simulator block reports mean scores across \projectname{}100, \emph{Household30}, and \emph{Workplace70}, using the strict \emph{Code-Rebuilt Score} $S_{\text{rebuilt}}$ for coding agents and compiled-scene scores for specialized baselines marked with slashes under \emph{Harness}.
  Best and second-best scores per column are \textbf{bolded} and \underline{underlined}, and component-level breakdowns appear in Supplementary Sec.~\ref{sec:supp-results}.}
  \label{tab:main-results}
  \footnotesize
  \riftColSep{0pt}
  \renewcommand{\arraystretch}{1.18}
  \begin{tabular*}{\textwidth}{@{\extracolsep{\fill}}lc@{\hspace{8pt}} ccc@{\hspace{9pt}} ccc@{\hspace{9pt}} ccc@{}}
    \toprule
    \multirow{2}{*}{\textbf{Model / Method}} & \multirow{2}{*}{\textbf{Harness}}
    & \multicolumn{3}{c@{\hspace{9pt}}}{\textbf{MuJoCo}~\uparr} & \multicolumn{3}{c@{\hspace{9pt}}}{\textbf{Isaac Sim}~\uparr} & \multicolumn{3}{c}{\textbf{SAPIEN}~\uparr} \\
    \cmidrule(r{9pt}){3-5} \cmidrule(r{9pt}){6-8} \cmidrule{9-11}
    &
    & {\scriptsize \textbf{Overall}} & {\scriptsize \textbf{Household}} & {\scriptsize \textbf{Workplace}}
    & {\scriptsize \textbf{Overall}} & {\scriptsize \textbf{Household}} & {\scriptsize \textbf{Workplace}}
    & {\scriptsize \textbf{Overall}} & {\scriptsize \textbf{Household}} & {\scriptsize \textbf{Workplace}} \\
    \midrule
    \rowcolor{rifttabgroup}
    \multicolumn{11}{@{}l}{\textbf{General-purpose coding agents}} \\
    Claude Opus 5.5 & Claude Code & \textbf{56.62} & \textbf{50.53} & \textbf{59.23} & \textbf{51.10} & \textbf{46.24} & \textbf{53.18} & \textbf{50.72} & \textbf{43.96} & \textbf{53.61} \\
    GPT-6 Astra & Codex & \underline{39.52} & \underline{23.36} & \underline{46.45} & \underline{41.57} & \underline{37.09} & \underline{43.49} & \underline{37.62} & \underline{25.32} & \underline{42.90} \\
    Kimi K3.1 Pro & Kimi Code & 26.29 & 21.98 & 28.14 & 23.55 & 18.76 & 25.60 & 26.01 & 21.51 & 27.94 \\
    DeepSeek-V4.1-Flash-1017 & DSH & 22.36 & 17.79 & 24.31 & 20.21 & 12.83 & 23.37 & 22.31 & 15.45 & 25.25 \\
    GLM 5.3 Flash & ZCode & 22.55 & 18.76 & 24.17 & 16.81 & 13.71 & 18.14 & 17.23 & 13.57 & 18.79 \\
    MiniMax M3.1-Flash & MiniMax Code & 18.76 & 19.10 & 18.62 & 11.43 & 10.05 & 12.03 & 16.37 & 17.39 & 15.94 \\
    \midrule
    \rowcolor{rifttabgroup}
    \multicolumn{11}{@{}l}{\textbf{Specialized 3D reconstruction methods}} \\
    SAM 3D Objects & / & 10.56 & 11.44 & 10.19 & 10.56 & 11.43 & 10.18 & 10.44 & 11.12 & 10.14 \\
    SAM 3D Objects + Particulate & / & 6.76 & 5.37 & 7.35 & 6.63 & 5.06 & 7.31 & 6.75 & 5.39 & 7.33 \\
    MIDI & / & 4.51 & 3.18 & 5.09 & 4.49 & 3.17 & 5.06 & 4.53 & 3.19 & 5.10 \\
    URDFormer & / & 2.60 & 1.32 & 3.14 & 2.61 & 1.32 & 3.17 & 2.51 & 1.16 & 3.09 \\
    \bottomrule
  \end{tabular*}

\end{table*}

\section{Experiments and Analysis}
\label{sec:experiments}

\subsection{Experimental Setup}
\label{sec:setup}
We evaluate the two system classes in Table~\ref{tab:main-results} across MuJoCo~\cite{todorov2012mujoco}, Isaac Sim~\cite{nvidia2025isaacsim}, and SAPIEN~\cite{xiang2020sapien}.
They comprise six frontier coding agents under their native harnesses alongside four specialized single-image 3D baselines: static rigid SAM 3D Objects~\cite{chen2025sam3d}, and articulated or multi-object SAM 3D Objects with Particulate~\cite{li2026particulate}, URDFormer~\cite{chen2024urdformer}, and MIDI~\cite{huang2025midi}.
Each specialized baseline takes only the single input image with official weights and default settings, passing through a shared segmentation and depth front end and unified physics exporter that aligns coordinates and metric scale without altering predicted geometry for fair scoring.

Each agent run receives the required GPU resources and a 240-minute active-time budget, saving workspace checkpoints every 15 minutes.
Main experiments and failure diagnostics evaluate all 100 tasks of \projectname{}100, while ablations use the 30-task \emph{Ablation30} with three tasks per category.
Main tables report the stricter \emph{Code-Rebuilt Score} $S_{\text{rebuilt}}$ for coding agents alongside compiled scene scores for specialized baselines without build scripts, with exact versions and configurations in Supplementary Sec.~\ref{sec:supp-methods}.

\subsection{Main Results}
\label{sec:main-results}
Table~\ref{tab:main-results} shows that single-image interactive reconstruction behaves fundamentally as feedback-driven constraint satisfaction rather than single-pass perception.
Equipped with in-container rendering inspection and physics checks, general-purpose coding agents outperform specialized baselines across all three simulators, as Claude Opus 5.5 reaches 50.72 to 56.62 and exceeds the best specialized baseline SAM 3D Objects by \textbf{about fivefold} on the three-simulator mean of 52.81 versus 10.52, ahead of GPT-6 Astra and the remaining four agents.
Conversely, Particulate, MIDI, and URDFormer consistently trail static SAM 3D Objects because open-loop part splitting and assembly compound physical mismatches, while top scores barely above 50 confirm that single-image interactive reconstruction remains far from solved.

Unlike specialized baselines that remain invariant across engines through a unified export pipeline, agents authoring native simulator code are directly constrained by low-level interface friction.
Five of six agents peak on MuJoCo with its clean declarative hierarchy and lose 2.2 to 7.3 points on the verbose OpenUSD schemas of Isaac Sim or imperative assembly APIs of SAPIEN, whereas GPT-6 Astra peaks at 41.57 on Isaac Sim yet still drops to 37.62 on SAPIEN.

Across scene morphologies, the two paradigms exhibit opposite domain preferences.
Leading coding agents score substantially higher on the orthogonal, modular workspaces of \emph{Workplace70} than on cluttered \emph{Household30}, whereas per-object prior-based SAM 3D Objects edges ahead on \emph{Household30}, highlighting natural complementarity between symbolic code compositionality and learned 3D shape priors.

Viewed alongside construction cost, closed-loop coding agents trade 1.5 to 2.6 active hours and 59k to 330k output tokens per task for interactive fidelity compared with seconds for feed-forward inference.
Yet higher test-time compute does not guarantee higher quality: Claude Opus 5.5 more than doubles the scores of DeepSeek-V4.1-Flash-1017 and GLM 5.3 Flash while using less active time and fewer output tokens.
Accurately translating visual and physical discrepancies into targeted code edits is thus the decisive watershed.

\begin{figure*}[t]
  \centering
  \includegraphics[pagebox=cropbox,width=\textwidth]{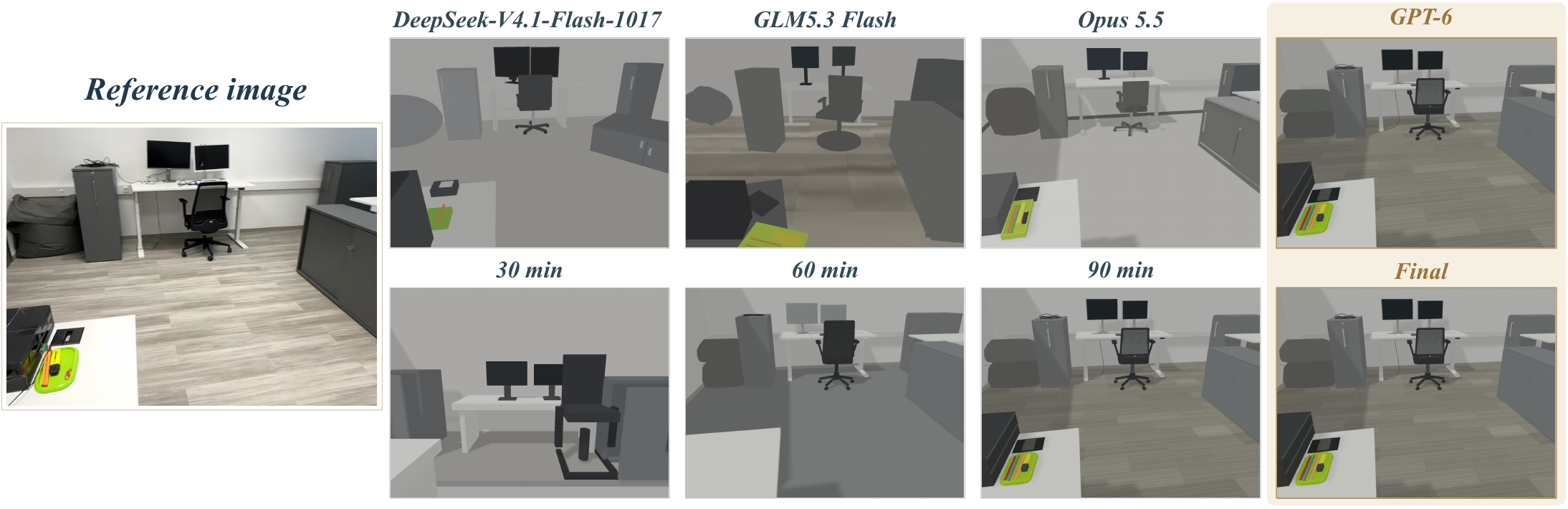}
  \caption{\textbf{Qualitative model comparison and checkpoint progression.} Final reconstructions across four coding agents appear in the top row, and intermediate 30-, 60-, and 90-minute checkpoints for GPT-6 Astra appear below up to the shared final scene boxed on the right.}
  \label{fig:qualitative}
\end{figure*}
Figure~\ref{fig:qualitative} shows this gap on an office photograph.
DeepSeek-V4.1-Flash-1017 and GLM 5.3 Flash produce coarse block assemblies at the wrong scale, with misplaced chairs and incomplete cabinets.
Claude Opus 5.5 and GPT-6 Astra recover the room layout, the desk with its monitors, the office chairs, and the low cabinets along the walls.
The checkpoints of GPT-6 Astra show how such a scene is built.
By 30\,min, the agent has a loadable room shell and a coarse desk layout.
At 60\,min and 90\,min, visual and physical checks guide it to refine the chairs, monitors, cabinets, and handles.

\subsection{Controlled Ablations}
\label{sec:ablations}
\begin{table}[t]
  \centering
  \caption{\textbf{Harness ablations on \emph{Ablation30}.}
  Each entry reports the mean \emph{Code-Rebuilt Score} $S_{\text{rebuilt}}$ across 30 stratified tasks under each model's native harness, the shared Claude Code harness, and the Bash-only \emph{mini-SWE-agent} loop; the highest mean within each model block is \textbf{bolded}.}
  \label{tab:ablations}
  \scriptsize
  \riftColSep{1pt}
  \renewcommand{\arraystretch}{1.15}
  \begin{tabularx}{\columnwidth}{@{}ll*{4}{>{\centering\arraybackslash}X}@{}}
    \toprule
    \textbf{Model} & \textbf{Harness} & {\fontsize{5.6pt}{6.6pt}\selectfont\textbf{MuJoCo}\uparr} & {\fontsize{5.6pt}{6.6pt}\selectfont\textbf{Isaac}\uparr} & {\fontsize{5.6pt}{6.6pt}\selectfont\textbf{SAPIEN}\uparr} & {\fontsize{5.6pt}{6.6pt}\selectfont\textbf{Mean}\uparr} \\
    \midrule
    \rowcolor{rifttabhi}
    Claude Opus 5.5 & Claude Code & 58.26 & 56.03 & 54.07 & \textbf{56.12} \\
    \cmidrule(l{2pt}r{2pt}){1-6}
    \riftGroupFirst{3}{GPT-6 Astra} & Codex & 39.44 & 41.54 & 38.14 & 39.71 \\
     & Claude Code & 43.26 & 43.94 & 44.28 & \textbf{43.82} \\
    \riftGroupLast{3}{GPT-6 Astra} & mini-SWE-agent & 41.46 & 44.58 & 41.44 & 42.49 \\
    \cmidrule(l{2pt}r{2pt}){1-6}
    \riftGroupFirst{3}{\riftTwoLine{DeepSeek-V4.1-}{Flash-1017}} & DSH & 20.01 & 23.80 & 24.02 & 22.61 \\
     & Claude Code & 33.83 & 24.21 & 25.63 & \textbf{27.89} \\
    \riftGroupLast{3}{\riftTwoLine{DeepSeek-V4.1-}{Flash-1017}} & mini-SWE-agent & 27.95 & 25.57 & 22.58 & 25.37 \\
    \cmidrule(l{2pt}r{2pt}){1-6}
    \riftGroupFirst{3}{GLM 5.3 Flash} & ZCode & 23.43 & 17.59 & 19.58 & 20.20 \\
     & Claude Code & 20.60 & 17.53 & 15.80 & 17.97 \\
    \riftGroupLast{3}{GLM 5.3 Flash} & mini-SWE-agent & 26.31 & 26.74 & 14.20 & \textbf{22.42} \\
    \bottomrule
  \end{tabularx}
\end{table}

\begin{table}[t]
  \centering
  \caption{\textbf{Reasoning-effort ablation on \emph{Ablation30}.} Three-simulator mean \emph{Code-Rebuilt Score} $S_{\text{rebuilt}}$ under each native CLI harness across supported effort levels, with the default \emph{High} setting shaded, the best level per model \textbf{bolded}, and dashes marking unsupported levels.}
  \label{tab:effort-ablation}
  \scriptsize
  \riftColSep{0.8pt}
  \renewcommand{\arraystretch}{1.25}
  \setlength{\aboverulesep}{0pt}\setlength{\belowrulesep}{0pt}\setlength{\extrarowheight}{1.6pt}
  \begin{tabularx}{\columnwidth}{@{}l>{\hsize=1.2\hsize\centering\arraybackslash}X>{\hsize=0.9\hsize\centering\arraybackslash}X>{\hsize=1.2\hsize\centering\arraybackslash}X>{\columncolor{rifttabhi}\hsize=1.0\hsize\centering\arraybackslash}X>{\hsize=1.0\hsize\centering\arraybackslash}X>{\hsize=0.85\hsize\centering\arraybackslash}X>{\hsize=0.85\hsize\centering\arraybackslash}X@{}}
    \toprule
    \textbf{Model} & \textbf{Minimal} & \textbf{Low} & \textbf{Medium} & \textbf{High} & \textbf{XHigh} & \textbf{Max} & \textbf{Ultra} \\
    \midrule
    GPT-6 Astra   & --    & 37.19 & 36.39 & 39.71 & 46.46 & 47.78 & \textbf{48.90} \\
    \riftTwoLine{DeepSeek-V4.1-}{Flash-1017} & 21.63 & 23.10 & 22.16 & 22.61 & 22.32 & \textbf{24.91} & -- \\
    GLM 5.3 Flash & --    & 12.41 & --    & \textbf{20.20} & -- & 12.97 & -- \\
    \bottomrule
  \end{tabularx}
\end{table}

Each agent in Table~\ref{tab:main-results} runs under its native harness, so its score mixes model capability with harness design.
Table~\ref{tab:ablations} separates the two on \emph{Ablation30} with two controls.
The first runs the other models under Claude Code, the harness of the leading model.
The second uses \emph{mini-SWE-agent}~\cite{minisweagent2025,yang2024sweagent}, a minimal loop that gives the model a single Bash tool and an uncompressed conversation history.
Supplementary Table~\ref{tab:supp-agent-configs} lists the settings of every harness.
Under Claude Code, the other three models average \textbf{17.97 to 43.82} across simulators, remaining well below the \textbf{56.12} of Claude Opus 5.5.
Even when stripped down to a single Bash tool in \emph{mini-SWE-agent}, GPT-6 Astra, DeepSeek-V4.1-Flash-1017, and GLM 5.3 Flash still reach 42.49, 25.37, and 22.42 without performance collapse.
The paradigm advantage and the lead of Claude Opus 5.5 therefore stem mainly from intrinsic 3D perception and error correction rather than outer scaffolding.

Harness design still shapes how models exercise that capability.
Structured tools and frequent checks in Claude Code lift GPT-6 Astra from 39.71 to 43.82 and DeepSeek-V4.1-Flash-1017 from 22.61 to 27.89 across all three simulators to their highest overall means, yet lower GLM 5.3 Flash from 20.20 to 17.97 on every engine.
Meanwhile, retaining uncompacted history in \emph{mini-SWE-agent} edges past native harnesses on MuJoCo and Isaac Sim, whereas verbose imperative scripts on SAPIEN pull GLM 5.3 Flash and DeepSeek-V4.1-Flash-1017 down from 19.58 and 24.02 to 14.20 and 22.58.

Table~\ref{tab:effort-ablation} varies the reasoning effort per turn, with per-simulator scores in Supplementary Table~\ref{tab:supp-effort-by-sim}.
More reasoning does not help every model.
GPT-6 Astra rises from 36.39 and 37.19 at \emph{Medium} and \emph{Low} through 39.71 at \emph{High} to 46.46 through 48.90 at \emph{XHigh} through \emph{Ultra}.
DeepSeek-V4.1-Flash-1017 stays between 21.63 and 24.91 across all levels.
GLM 5.3 Flash peaks at 20.20 under \emph{High} and falls to 12.97 at \emph{Max}, where long reasoning in each turn leaves fewer turns for refinement.
Supplementary Sec.~\ref{sec:supp-comparisons} reports resource usage and quality over time.

\subsection{Failure Diagnosis}
\label{sec:failure-diagnosis}
\begin{figure}[t]
  \centering
  \includegraphics[width=\columnwidth]{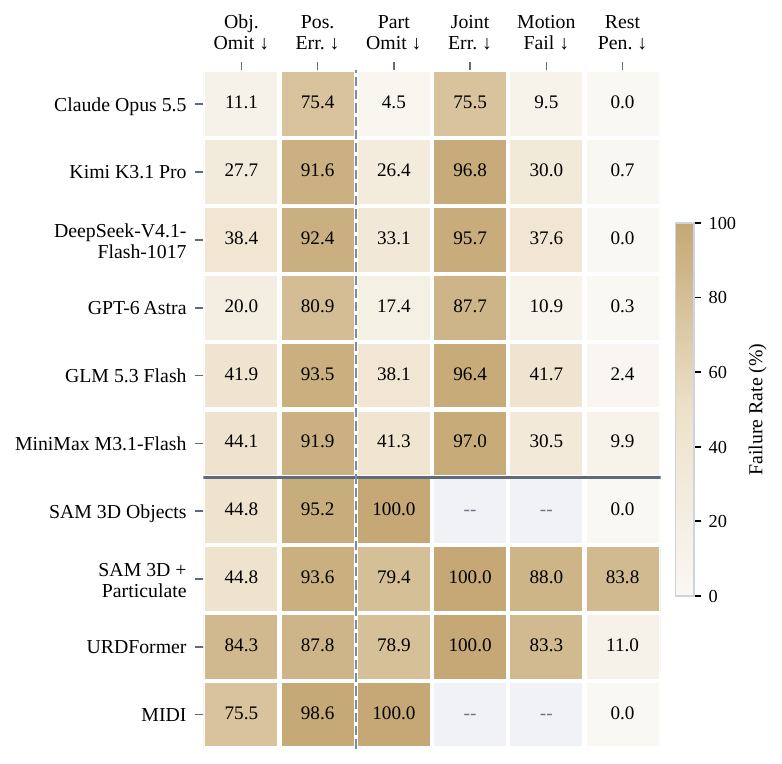}
  \caption{\textbf{Failure diagnostics averaged across simulators.}
  Each cell reports the mean percentage failure rate across MuJoCo, Isaac Sim, and SAPIEN on frozen submissions across six metrics: object omission \emph{Obj.\ Omit}, matched-object position error \emph{Pos.\ Err.}, part omission \emph{Part Omit}, matched-part joint error \emph{Joint Err.}, actuation failure \emph{Motion Fail}, and static rest interpenetration \emph{Rest Pen.}
  Dashes denote metrics not applicable to static baselines, and per-simulator breakdowns appear in Supplementary Table~\ref{tab:supp-failure-diagnostics}.}
  \label{fig:failure-diagnostics-heatmap}
\end{figure}
Deterministic clean-environment code reproduction is no longer the primary bottleneck limiting top agents near 50 points, isolating the core deficit to 3D geometric and physical modeling.
Across Table~\ref{tab:main-results} and Supplementary Table~\ref{tab:supp-failure-diagnostics}, four of six coding agents achieve zero invalid submissions and zero offline rebuild failures across all engines, while DeepSeek-V4.1-Flash-1017 incurs 2.0\% to 6.0\% rebuild failures, MiniMax M3.1-Flash incurs 5.0\% to 15.0\% invalid or unrebuildable runs, and $S_{\text{frozen}} - S_{\text{rebuilt}}$ gaps stay within 1 point for every model.

Figure~\ref{fig:failure-diagnostics-heatmap} exposes a sharp divide between semantic instantiation and metric 3D grounding.
Claude Opus 5.5 lowers object and part omission to 11.1\% and 4.5\% versus 44.8\% to 100.0\% on specialized baselines.
On matched elements, however, position and joint error rates remain universally high at \textbf{75.4\% to 100.0\%} across all systems, reflecting both multimodal limits in lifting 2D pixels to metric 3D poses and axes and the inherent single-view ambiguity of cabinet depth, hinge side, and internal slides behind closed doors.

Static geometric consistency does not guarantee dynamic operability.
Through in-container settling checks, most coding agents keep static rest penetration between 0.0\% and 2.4\% compared with 83.8\% on open-loop Particulate, yet under controlled torque actuation, \textbf{over 80\%} of matched parts in specialized articulated baselines jam, non-Opus coding agents suffer 10.9\% to 41.7\% motion failure, and even Claude Opus 5.5 reaches 19.5\% on Isaac Sim.
Minute door-frame clearance collisions, off-axis hinges, and inverted joint limits remain dormant at rest and surface as collider self-locking only during motion, foreshadowing stricter failures under physical robot contact.

\begin{figure}[t]
  \centering
  \includegraphics[width=\columnwidth]{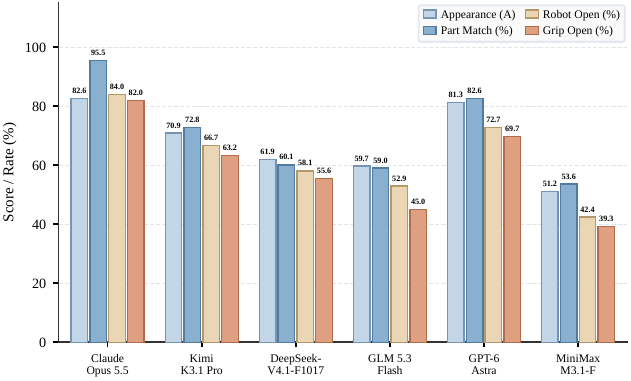}
  \caption{\textbf{Visual appearance vs.\ simulated robot manipulation.}
  Three-simulator means on \projectname{}100 under clean rebuilds for the six evaluated coding agents, comparing \emph{Appearance} $A$ against \emph{Part Match}, total Franka Panda opening rate \emph{Robot Open} across 323 ground-truth parts, and pure gripper-only opening rate \emph{Grip Open} without attach fallback.}
  \label{fig:robot-manipulation}
\end{figure}
\begin{figure*}[t]
  \centering
  \includegraphics[pagebox=cropbox,width=\textwidth]{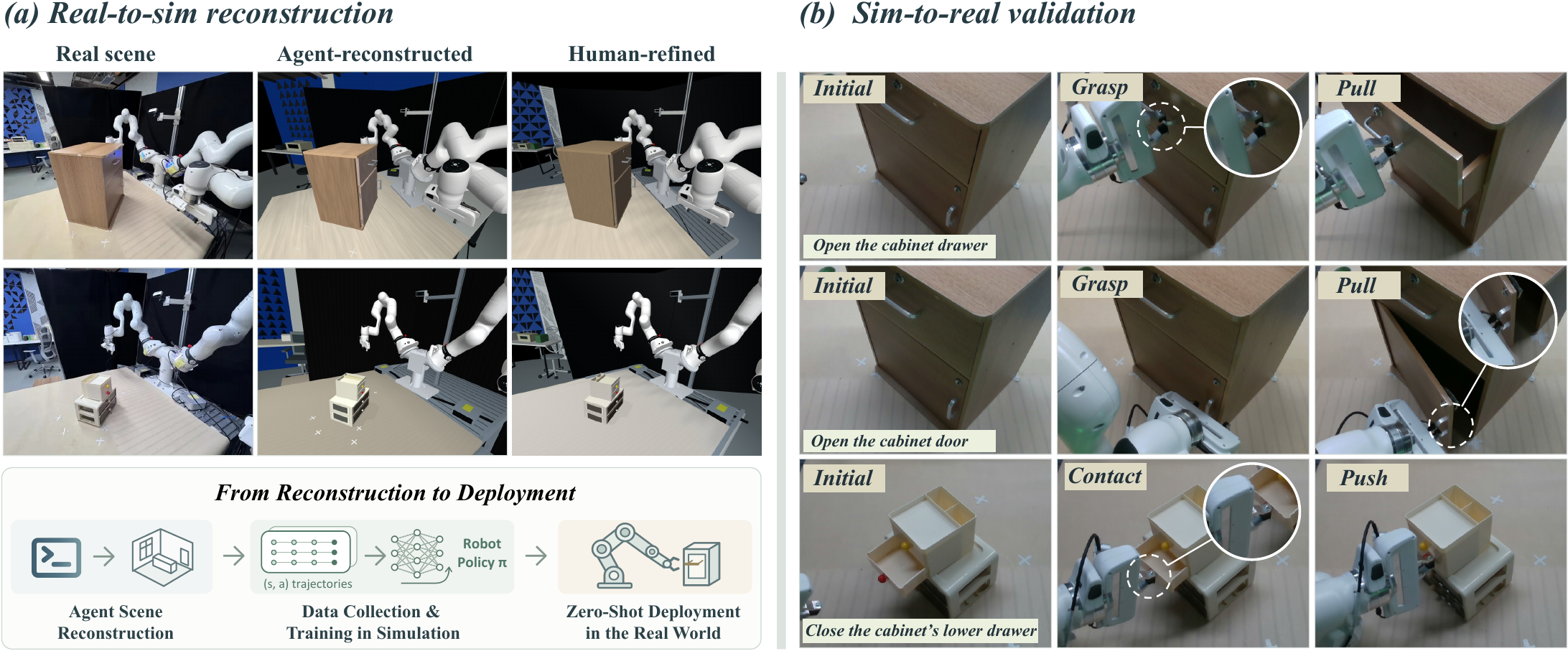}
  \caption{\textbf{Real-to-sim-to-real validation on a physical bimanual Franka platform.}
  \textbf{Panel a:} External reference photographs alongside raw \emph{Agent-reconstructed} MuJoCo scenes built by GPT-6 Astra with Codex, \emph{Human-refined} counterparts, and the end-to-end deployment pipeline. Third-person views are shown solely for visual comparison.
  \textbf{Panel b:} Zero-shot real-robot rollouts across three contact tasks: opening the wooden cabinet's upper drawer and lower hinged door, and closing the organizer's lower drawer.}
  \label{fig:real-to-sim-to-real}
\end{figure*}
\subsection{Visual Fidelity vs.\ Embodied Operability}
\label{sec:visual-vs-interaction}
Visual appearance fidelity substantially overestimates embodied robot operability, as convincing 2D renders alone fail to support reliable physical manipulation.
Figure~\ref{fig:robot-manipulation} juxtaposes 2D appearance scores against simulated Franka Panda opening rates across all 323 ground-truth parts.
Although Kimi K3.1 Pro, DeepSeek-V4.1-Flash-1017, and GLM 5.3 Flash achieve appearance scores of 70.9, 61.9, and 59.7, all six evaluated agents suffer a consistent two-stage drop from \emph{Part Match} to \emph{Robot Open} and \emph{Grip Open}, falling even on Claude Opus 5.5 from \textbf{95.5\% to 84.0\% and 82.0\%}.

Simulated robot manipulation further isolates two hidden physical failure layers missed by both 2D rendering and direct joint-torque checks, namely macro workspace reachability and micro handle contact geometry.
The first drop from \emph{Part Match} to \emph{Robot Open} reflects not only internal joint jamming but also nearby furniture obstructing arm base placement and off-axis arcs extending beyond arm reachability.
Removing kinematic attach fallback reveals a further drop to pure parallel-jaw \emph{Grip Open}, such as from 52.9\% to 45.0\% on GLM 5.3 Flash.
This gap exposes handle slippage caused by missing, shallowly recessed, or overly thin pulls, anticipating the alignment errors and limit stops observed during real-world transfer in Sec.~\ref{sec:real-world}.

\subsection{Agreement with Human Judgment}
\label{sec:human-study}
\begin{table}[t]
  \centering
  \caption{\textbf{Blind pairwise human preferences and matched automated \projectname{} scores across six coding agents.}
  Each model receives 100 votes per question across 150 pairs on 10 MuJoCo tasks, alongside mean $S_{\text{rebuilt}}$ on the identical 60 submissions.}
  \label{tab:human-study}
  \scriptsize
  \riftColSep{0pt}
  \renewcommand{\arraystretch}{1.35}
  \begin{tabularx}{\columnwidth}{@{}l>{\hsize=1.10\hsize\linewidth=\hsize\centering\arraybackslash}X>{\hsize=0.92\hsize\linewidth=\hsize\centering\arraybackslash}X>{\hsize=1.10\hsize\linewidth=\hsize\centering\arraybackslash}X>{\hsize=0.92\hsize\linewidth=\hsize\centering\arraybackslash}X>{\hsize=0.96\hsize\linewidth=\hsize\centering\arraybackslash}X@{}}
    \toprule
    & \multicolumn{2}{c}{\textbf{Q1: Scene}} & \multicolumn{2}{c}{\textbf{Q2: Robot}} & \textbf{Auto.} \\
    \cmidrule(lr{3pt}){2-3} \cmidrule(lr{3pt}){4-5} \cmidrule(l{2pt}){6-6}
    \textbf{Model} & \textbf{W/L/T} & \textbf{Win\%}\uparr & \textbf{W/L/T} & \textbf{Win\%}\uparr & \textbf{\projectname{}}\uparr \\
    \midrule
    \rowcolor{rifttabalt}Claude Opus 5.5 & 75/6/19 & \textbf{92.59} & 48/19/33 & \underline{71.64} & \textbf{66.22} \\
    GPT-6 Astra & 66/9/25 & \underline{88.00} & 52/5/43 & \textbf{91.23} & \underline{42.01} \\
    \rowcolor{rifttabalt}Kimi K3.1 Pro & 49/25/26 & 66.22 & 36/26/38 & 58.06 & 29.84 \\
    \riftTwoLine{DeepSeek-V4.1-}{Flash-1017} & 26/49/25 & 34.67 & 18/38/44 & 32.14 & 28.34 \\
    \rowcolor{rifttabalt}GLM 5.3 Flash & 9/69/22 & 11.54 & 12/47/41 & 20.34 & 22.35 \\
    MiniMax M3.1-Flash & 4/71/25 & 5.33 & 10/41/49 & 19.61 & 12.81 \\
    \bottomrule
  \end{tabularx}
\end{table}

We check whether \projectname{} scores agree with human judgment through a double-blind pairwise study.
The study covers all six native coding agents on 10 MuJoCo tasks, giving 15 model pairs per photograph and 150 pairs in total.
Five raters each judge 60 pairs in an offline browser interface, so every pair receives two independent judgments and the study collects 300 comparisons.
Each submission appears as an anonymized evidence sheet with six rows: camera overlays, collision geometry, top-down layouts with parts closed and open, per-part articulation, an oblique overview, and simulated Franka Panda rollouts.
Left and right positions are randomized for each rater.
Raters vote win, loss, or tie on two questions, \emph{Scene Reconstruction Quality} as Q1 and \emph{Robot Manipulation Readiness} as Q2, and we rank models by their win rate over non-tied votes.

Table~\ref{tab:human-study} compares the votes with \projectname{} scores on the same 60 submissions.
For Q1, the two rankings are identical.
\projectname{} scores fall from 66.22 for Claude Opus 5.5 to 12.81 for MiniMax M3.1-Flash, and human win rates fall from 92.59\% to 5.33\% in the same order, giving Spearman $\rho=1.00$ and Kendall $\tau=1.00$ at $p=0.0028$.
For Q2, the rank correlation is $\rho=0.94$ at $p=0.0048$.
Here GPT-6 Astra ranks above Claude Opus 5.5 with win rates of 91.23\% and 71.64\%, which matches its higher simulated \emph{Robot Open} rate of 93.8\% versus 68.8\% on the 16 scored parts.
Raters also agree less with each other on Q2.
Exact agreement drops from 69.3\% to 48.0\%, and $\kappa$ drops from $[0.396,0.686]$ to $[-0.147,0.384]$.
Sub-centimeter clearances and grasp contact are hard to judge from renders, whereas \projectname{} measures them directly in simulation.
Supplementary Sec.~\ref{sec:supp-human-study} gives the full protocol and statistics.

\subsection{Real-World Robot Validation}
\label{sec:real-world}
As shown in Figure~\ref{fig:real-to-sim-to-real}, we conduct controlled trials on a bimanual Franka Research 3 platform across two real cabinets and three contact tasks to test whether single-image agent reconstructions can support zero-shot sim-to-real transfer of a vision-language-action model.
Without depth maps or camera calibration, GPT-6 Astra under Codex receives only a single external image, the robot model, and outer tabletop and cabinet dimensions to anchor metric scale.
\begin{table}[t]
  \centering
  \caption{\textbf{Simulation vs.\ zero-shot real-robot manipulation.}
  Success rates of the deployed 10,000-step $\pi_{0.5}$ checkpoints in their training scenes with 100 seeds per task and on the physical bimanual Franka Research 3 platform with 10 trials per task.}
  \label{tab:sim2real}
  \scriptsize
  \riftColSep{2pt}
  \renewcommand{\arraystretch}{1.16}
  \begin{tabularx}{\columnwidth}{@{}l*{4}{>{\centering\arraybackslash}X}@{}}
    \toprule
    \multirow{2}{*}{\textbf{Policy}}
    & \textbf{Open} & \textbf{Open} & \textbf{Close} & \textbf{Mean} \\
    & \textbf{Drawer}\,\uparr & \textbf{Door}\,\uparr & \textbf{Drawer}\,\uparr & \textbf{\%}\,\uparr \\
    \midrule
    \multicolumn{5}{@{}>{\columncolor{rifttabgroup}}p{\columnwidth}@{}}{%
      \textbf{In-domain simulation, \% over 100 seeds per task}} \\
    $\pi_{0.5}$ + \emph{Agent-recon.} & \textbf{100.0} & \textbf{43.0} & \textbf{37.0} & \textbf{60.0} \\
    $\pi_{0.5}$ + \emph{Human-refined} & 82.0 & 38.0 & 31.0 & 50.3 \\
    \midrule
    \multicolumn{5}{@{}>{\columncolor{rifttabgroup}}p{\columnwidth}@{}}{%
      \textbf{Zero-shot real-robot, successes of 10 trials and mean \%}} \\
    Unfinetuned $\pi_{0.5}$ & 0/10 & 0/10 & 0/10 & 0.0 \\
    $\pi_{0.5}$ + \emph{Agent-recon.} & 2/10 & 1/10 & \textbf{2/10} & 16.7 \\
    $\pi_{0.5}$ + \emph{Human-refined} & \textbf{3/10} & \textbf{4/10} & \textbf{2/10} & \textbf{30.0} \\
    \bottomrule
  \end{tabularx}
\end{table}

From this input, the agent builds the raw \emph{Agent-reconstructed} scene, paired with a \emph{Human-refined} counterpart calibrated in placement, interpenetration, local geometry, and appearance.
After freezing both scenes and attaching three calibrated onboard cameras, a scripted inverse-kinematics planner collects 300 valid trajectories per scene under paired random perturbations.
We then separately fine-tune the official $\pi_{0.5}$~\cite{intelligence2025pi05} base model under identical settings, with full hardware and training details in Supplementary Sec.~\ref{sec:supp-real-world}.

As shown in Table~\ref{tab:sim2real}, policies trained on \emph{Agent-reconstructed} and \emph{Human-refined} both converge most evenly in held-out simulation at the 10,000-step checkpoint, averaging 60.0\% and 50.3\% within their respective virtual environments.
Deployed alongside the unfinetuned $\pi_{0.5}$ baseline across 10 real-robot trials per task, the three policies achieve mean success rates of \textbf{0.0\%}, \textbf{16.7\%} with 5 of 30 successes, and \textbf{30.0\%} with 9 of 30, providing initial evidence that data synthesized from single-image agent reconstruction can support zero-shot real-robot contact manipulation.
Notably, multi-dimensional calibration lifts hinged door opening from 1/10 to 4/10 and drawer opening from 2/10 to 3/10 while holding lower-drawer closing at 2/10, as hinged doors are most sensitive to axis and handle alignment.
Physical trials invert the simulation ranking because in-domain evaluation rewards fitting each scene's own virtual deviations, whereas physical deployment directly tests real-world alignment.
Because the policy predicts joint targets directly, single-view offsets in placement, geometry, and appearance shift simulated trajectories from real onboard views and frequently trigger arm joint-limit stops on the raw-scene policy, whereas multi-dimensional calibration markedly reduces limit violations and leaves mainly local handle-alignment errors. This finding confirms that reliable sim-to-real transfer requires joint geometric, kinematic, and visual fidelity beyond 2D appearance.

\section{Conclusion}
\label{sec:conclusion}
We presented \projectname{}, a cross-simulator benchmark pairing laser-scanned ground truth with isolated containers to evaluate single-image interactive indoor reconstruction across coding agents and specialized 3D baselines.
Our experiments show that the strongest closed-loop coding agent outperforms the best single-pass baseline by about fivefold through intrinsic 3D perception and iterative error correction, an advantage that persists under a Bash-only harness, while complementing learned shape priors, yet coding agents remain bottlenecked by metric 3D grounding, dynamic collider self-locking, and contact geometry.
Automated \projectname{} rankings match blind human judgments of scene quality while capturing contact failures that human raters struggle to identify from renders.
Real-robot trials on a bimanual Franka platform further demonstrate viable zero-shot sim-to-real transfer from single-image reconstructions and underscore multi-dimensional fidelity evaluation.

A primary limitation of our study is that a single static photograph with closed doors leaves occluded cavity depth, hinge side, and internal slide travel inherently ambiguous from one view alone.
Future work will incorporate sparse casual multi-view images to resolve occlusion uncertainty, fuse programmatic code generation with learned 3D shape priors, and extend sim-to-real policy transfer across broader indoor scenes and manipulation tasks.

{
    \footnotesize
    \bibliographystyle{ieeenat_fullname}
    \bibliography{main}

\begin{thebibliography}{37}
\providecommand{\natexlab}[1]{#1}
\providecommand{\url}[1]{\texttt{#1}}
\expandafter\ifx\csname urlstyle\endcsname\relax
  \providecommand{\doi}[1]{doi: #1}\else
  \providecommand{\doi}{doi: \begingroup \urlstyle{rm}\Url}\fi

\bibitem[Baruch et~al.(2021)Baruch, Chen, Dehghan, Dimry, Feigin, Fu, Gebauer,
  Joffe, Kurz, Schwartz, and Shulman]{baruch2021arkitscenes}
Gilad Baruch, Zhuoyuan Chen, Afshin Dehghan, Tal Dimry, Yuri Feigin, Peter Fu,
  Thomas Gebauer, Brandon Joffe, Daniel Kurz, Arik Schwartz, and Elad Shulman.
\newblock {ARKitScenes}: A diverse real-world dataset for {3D} indoor scene
  understanding using mobile {RGB-D} data.
\newblock In \emph{Proceedings of the Neural Information Processing Systems
  Track on Datasets and Benchmarks}, 2021.

\bibitem[Black et~al.(2025)Black, Brown, Darpinian, Dhabalia, Driess, Esmail,
  Equi, Finn, Fusai, Galliker, Ghosh, Groom, Hausman, Ichter, Jakubczak, Jones,
  Ke, LeBlanc, Levine, Li-Bell, Mothukuri, Nair, Pertsch, Ren, Shi, Smith,
  Springenberg, Stachowicz, Tanner, Vuong, Walke, Walling, Wang, Yu, and
  Zhilinsky]{intelligence2025pi05}
Kevin Black, Noah Brown, James Darpinian, Karan Dhabalia, Danny Driess, Adnan
  Esmail, Michael Equi, Chelsea Finn, Niccolo Fusai, Manuel~Y. Galliker, Dibya
  Ghosh, Lachy Groom, Karol Hausman, Brian Ichter, Szymon Jakubczak, Tim Jones,
  Liyiming Ke, Devin LeBlanc, Sergey Levine, Adrian Li-Bell, Mohith Mothukuri,
  Suraj Nair, Karl Pertsch, Allen~Z. Ren, Lucy~Xiaoyang Shi, Laura Smith,
  Jost~Tobias Springenberg, Kyle Stachowicz, James Tanner, Quan Vuong, Homer
  Walke, Anna Walling, Haohuan Wang, Lili Yu, and Ury Zhilinsky.
\newblock {$\pi_{0.5}$: A Vision-Language-Action Model with Open-World
  Generalization}.
\newblock \emph{arXiv preprint arXiv:2504.16054}, 2025.

\bibitem[Cai et~al.(2026)Cai, Zhu, Shang, Li, Pollefeys, Wu, Vedaldi, and
  Zheng]{cai2026artisan}
Xudong Cai, Ruijie Zhu, Haoliang Shang, Ruining Li, Marc Pollefeys, Shangzhe
  Wu, Andrea Vedaldi, and Chuanxia Zheng.
\newblock {Towards Physically Grounded Articulated 3D Assets: A Multi-Agent
  Pipeline and Benchmark}.
\newblock \emph{arXiv preprint arXiv:2610.13947}, 2026.

\bibitem[Chen et~al.(2025)Chen, Jiang, Liu, Gupta, Li, Zhao, and
  Wang]{chen2025physgen3d}
Boyuan Chen, Hanxiao Jiang, Shaowei Liu, Saurabh Gupta, Yunzhu Li, Hao Zhao,
  and Shenlong Wang.
\newblock {PhysGen3D}: Crafting a miniature interactive world from a single
  image.
\newblock In \emph{Proceedings of the IEEE/CVF Conference on Computer Vision
  and Pattern Recognition}, pages 6178--6189, 2025.

\bibitem[Chen et~al.(2026)Chen, Xia, Peng, Zhang, Jing, Ma, Qian, Cheng, Jiao,
  Zhou, Qu, Ye, Zhang, Wang, Zeng, Chen, Yang, Zeng, Luo, Wang, Liu, Yuille,
  Shi, Zheng, Li, Jiang, and Chen]{chen2026agenticreal2sim}
Guanxiong Chen, Qianjun Xia, Jiawei Peng, Heng Zhang, Pengyu Jing, Bole Ma,
  Justin Qian, Yixian Cheng, Ziyi Jiao, Bingyang Zhou, Yiduo Qu, Luoxin Ye,
  Kaifeng Zhang, Kunyi Wang, Weijia Zeng, Yunuo Chen, Pengzhi Yang, Ziqiu Zeng,
  Siyuan Luo, Huamin Wang, Chao Liu, Alan Yuille, Fan Shi, Changxi Zheng,
  Yunzhu Li, Chenfanfu Jiang, and Peter~Yichen Chen.
\newblock {Agentic Real2Sim}: Physics-based world modeling with vision-language
  agents.
\newblock \emph{arXiv preprint arXiv:2607.19190}, 2026.

\bibitem[Chen et~al.(2024)Chen, Walsman, Memmel, Mo, Fang, Vemuri, Wu, Fox, and
  Gupta]{chen2024urdformer}
Zoey Chen, Aaron Walsman, Marius Memmel, Kaichun Mo, Alex Fang, Karthikeya
  Vemuri, Alan Wu, Dieter Fox, and Abhishek Gupta.
\newblock {URDFormer}: A pipeline for constructing articulated simulation
  environments from real-world images.
\newblock In \emph{Robotics: Science and Systems}, 2024.

\bibitem[Dai et~al.(2024)Dai, Wong, Jiang, Wang, Gokmen, Zhang, Wu, and
  Li]{dai2024acdc}
Tianyuan Dai, Josiah Wong, Yunfan Jiang, Chen Wang, Cem Gokmen, Ruohan Zhang,
  Jiajun Wu, and Fei-Fei Li.
\newblock Automated creation of digital cousins for robust policy learning.
\newblock In \emph{Conference on Robot Learning}, 2024.

\bibitem[Delitzas et~al.(2024)Delitzas, Takmaz, Tombari, Sumner, Pollefeys, and
  Engelmann]{delitzas2024scenefun3d}
Alexandros Delitzas, Ayca Takmaz, Federico Tombari, Robert Sumner, Marc
  Pollefeys, and Francis Engelmann.
\newblock {SceneFun3D}: Fine-grained functionality and affordance understanding
  in {3D} scenes.
\newblock In \emph{Proceedings of the IEEE/CVF Conference on Computer Vision
  and Pattern Recognition}, pages 14531--14542, 2024.

\bibitem[Halacheva et~al.(2025)Halacheva, Miao, Zaech, Wang, Van~Gool, and
  Paudel]{halacheva2025articulate3d}
Anna-Maria Halacheva, Yang Miao, Jan-Nico Zaech, Xi Wang, Luc Van~Gool, and
  Danda~Pani Paudel.
\newblock {Articulate3D}: Holistic understanding of {3D} scenes as universal
  scene description.
\newblock In \emph{Proceedings of the IEEE/CVF International Conference on
  Computer Vision}, pages 5633--5644, 2025.

\bibitem[Hollis et~al.(2026)Hollis, Fan, and Yang]{hollis2026sceneagent}
Luke Hollis, Tianxing Fan, and Heng Yang.
\newblock {SceneAgent}: An agentic pipeline from {3D} and {4D} captures to
  simulation environments.
\newblock \emph{arXiv preprint arXiv:2610.34745}, 2026.

\bibitem[Hua et~al.(2024)Hua, Liu, Macaluso, Lin, Zhang, Xu, and
  Wang]{hua2024gensim2}
Pu Hua, Minghuan Liu, Annabella Macaluso, Yunfeng Lin, Weinan Zhang, Huazhe Xu,
  and Lirui Wang.
\newblock {GenSim2}: Scaling robot data generation with multi-modal and
  reasoning {LLMs}.
\newblock In \emph{Conference on Robot Learning}, 2024.

\bibitem[Huang et~al.(2025)Huang, Guo, An, Yang, Li, Zou, Liang, Liu, Cao, and
  Sheng]{huang2025midi}
Zehuan Huang, Yuan-Chen Guo, Xingqiao An, Yunhan Yang, Yangguang Li, Zi-Xin
  Zou, Ding Liang, Xihui Liu, Yan-Pei Cao, and Lu Sheng.
\newblock {MIDI}: Multi-instance diffusion for single image to {3D} scene
  generation.
\newblock In \emph{Proceedings of the IEEE/CVF Conference on Computer Vision
  and Pattern Recognition}, pages 23646--23657, 2025.

\bibitem[Jiang et~al.(2025)Jiang, Hsu, Zhang, Yu, Wang, and
  Li]{jiang2025phystwin}
Hanxiao Jiang, Hao-Yu Hsu, Kaifeng Zhang, Hsin-Ni Yu, Shenlong Wang, and Yunzhu
  Li.
\newblock {PhysTwin}: Physics-informed reconstruction and simulation of
  deformable objects from videos.
\newblock In \emph{Proceedings of the IEEE/CVF International Conference on
  Computer Vision}, pages 7219--7230, 2025.

\bibitem[Le et~al.(2025)Le, Xie, Liang, Wang, Yang, Ma, Vedder, Krishna,
  Jayaraman, and Eaton]{le2025articulateanything}
Long Le, Jason Xie, William Liang, Hung-Ju Wang, Yue Yang, Yecheng~Jason Ma,
  Kyle Vedder, Arjun Krishna, Dinesh Jayaraman, and Eric Eaton.
\newblock {Articulate-Anything}: Automatic modeling of articulated objects via
  a vision-language foundation model.
\newblock In \emph{International Conference on Learning Representations}, 2025.

\bibitem[Li et~al.(2026{\natexlab{a}})Li, Yao, Zhou, Zheng, Rupprecht, Lasenby,
  Wu, and Vedaldi]{li2026particulate}
Ruining Li, Yuxin Yao, Matt Zhou, Chuanxia Zheng, Christian Rupprecht, Joan
  Lasenby, Shangzhe Wu, and Andrea Vedaldi.
\newblock {Particulate}: Feed-forward {3D} object articulation.
\newblock In \emph{Proceedings of the IEEE/CVF Conference on Computer Vision
  and Pattern Recognition}, 2026{\natexlab{a}}.

\bibitem[Li et~al.(2026{\natexlab{b}})Li, Shi, Dong, Zhang, Xu, Lee, Chang,
  Xiang, Pan, and Jiang]{li2026legoanything}
Xirui Li, Peng Shi, Mingwen Dong, Sheng Zhang, Zhuoyan Xu, Dongkyu Lee,
  Shuaichen Chang, Yi Xiang, Lin Pan, and Jiarong Jiang.
\newblock {LEGO-Anything: Coding Agents for 3D Scene Reconstruction}.
\newblock \emph{arXiv preprint arXiv:2609.36380}, 2026{\natexlab{b}}.

\bibitem[Lin et~al.(2026)Lin, Liu, Li, Lam, Wang, Hong, Gao, Jiang, Chi, Zheng,
  Xiao, Yang, and Gao]{lin2026frontier3dbench}
Xinjie Lin, Qingle Liu, Wenyi Li, John Lam, Xiaoqiu Wang, Deyao Hong, Mingju
  Gao, Zheng Jiang, Yizhe Chi, Youjie Zheng, Calvin Xiao, Kaisen Yang, and
  Huan-ang Gao.
\newblock {Frontier3DBench: Benchmarking Coding Agents for 3D Creation across
  Objects, Scenes, and Interactive Open Worlds}.
\newblock \emph{arXiv preprint arXiv:2611.09017}, 2026.

\bibitem[Liu et~al.(2025)Liu, Iliash, Chang, Savva, and
  Mahdavi-Amiri]{liu2025singapo}
Jiayi Liu, Denys Iliash, Angel~X. Chang, Manolis Savva, and Ali Mahdavi-Amiri.
\newblock {SINGAPO}: Single image controlled generation of articulated parts in
  objects.
\newblock In \emph{International Conference on Learning Representations}, 2025.

\bibitem[Mandi et~al.(2025)Mandi, Weng, Bauer, and Song]{mandi2024real2code}
Zhao Mandi, Yijia Weng, Dominik Bauer, and Shuran Song.
\newblock {Real2Code}: Reconstruct articulated objects via code generation.
\newblock In \emph{International Conference on Learning Representations}, 2025.

\bibitem[{NVIDIA}(2025)]{nvidia2025isaacsim}
{NVIDIA}.
\newblock {NVIDIA Isaac Sim}.
\newblock \url{https://github.com/isaac-sim/IsaacSim}, 2025.

\bibitem[Raistrick et~al.(2024)Raistrick, Mei, Kayan, Yan, Zuo, Han, Wen,
  Parakh, Alexandropoulos, Lipson, Ma, and Deng]{raistrick2024infinigenindoors}
Alexander Raistrick, Lingjie Mei, Karhan Kayan, David Yan, Yiming Zuo, Beining
  Han, Hongyu Wen, Meenal Parakh, Stamatis Alexandropoulos, Lahav Lipson, Zeyu
  Ma, and Jia Deng.
\newblock {Infinigen Indoors}: Photorealistic indoor scenes using procedural
  generation.
\newblock In \emph{Proceedings of the IEEE/CVF Conference on Computer Vision
  and Pattern Recognition}, pages 21783--21794, 2024.

\bibitem[Ren et~al.(2024)Ren, Wang, Cai, Tuominen, Kannala, and
  Rahtu]{ren2024mushroom}
Xuqian Ren, Wenjia Wang, Dingding Cai, Tuuli Tuominen, Juho Kannala, and Esa
  Rahtu.
\newblock {MuSHRoom}: Multi-sensor hybrid room dataset for joint {3D}
  reconstruction and novel view synthesis.
\newblock In \emph{Proceedings of the IEEE/CVF Winter Conference on
  Applications of Computer Vision}, pages 4508--4517, 2024.

\bibitem[{SAM 3D Team} et~al.(2025){SAM 3D Team}, Chen, Chu, Gleize, Liang,
  Sax, Tang, Wang, Guo, Hardin, Li, Lin, Liu, Ma, Sagar, Song, Wang, Yang,
  Zhang, Doll{\'a}r, Gkioxari, Feiszli, and Malik]{chen2025sam3d}
{SAM 3D Team}, Xingyu Chen, Fu-Jen Chu, Pierre Gleize, Kevin~J. Liang,
  Alexander Sax, Hao Tang, Weiyao Wang, Michelle Guo, Thibaut Hardin, Xiang Li,
  Aohan Lin, Jiawei Liu, Ziqi Ma, Anushka Sagar, Bowen Song, Xiaodong Wang,
  Jianing Yang, Bowen Zhang, Piotr Doll{\'a}r, Georgia Gkioxari, Matt Feiszli,
  and Jitendra Malik.
\newblock {SAM 3D}: 3dfy anything in images.
\newblock \emph{arXiv preprint arXiv:2511.16624}, 2025.

\bibitem[Shao et~al.(2026)Shao, He, Fan, Guo, Wang, Ren, Zhang, Xu, Li, Wang,
  Lin, Liu, and Zhang]{shao2026physcodegym}
Shuai Shao, Jinchen He, Cunxin Fan, Zihan Guo, Shijian Wang, Qihan Ren,
  Kangning Zhang, Zunnan Xu, Haoyuan Li, Hao Wang, Jianghao Lin, Weiwen Liu,
  and Weinan Zhang.
\newblock {PhysCodeGym: Benchmarking Coding Agents for Physically Verifiable 4D
  Reconstruction}.
\newblock \emph{arXiv preprint arXiv:2610.17288}, 2026.

\bibitem[Shen et~al.(2026)Shen, Kova{\v{c}}i{\v{c}}, Kulits, Wang, Li,
  Tenenbaum, Yuille, Chen, and Wu]{shen2026fourdecodebench}
Ruihong Shen, {\v{Z}}iga Kova{\v{c}}i{\v{c}}, Peter Kulits, Xingrui Wang,
  Zizhang Li, Joshua~B. Tenenbaum, Alan Yuille, Jieneng Chen, and Jiajun Wu.
\newblock {4DCodeBench: Benchmarking Agents on Inverse Graphics of Dynamic
  Scenes}.
\newblock \emph{arXiv preprint arXiv:2610.03715}, 2026.

\bibitem[{SWE-agent Team}(2025)]{minisweagent2025}
{SWE-agent Team}.
\newblock {mini-SWE-agent}.
\newblock \url{https://github.com/SWE-agent/mini-swe-agent}, 2025.

\bibitem[Tang et~al.(2026{\natexlab{a}})Tang, Zhang, Yang, Yan, Zhou, Mao, Han,
  Cao, Xu, He, Fu, Feng, and Huang]{tang2026video2world}
Jinzhou Tang, Zijun Zhang, Jing Yang, Yuchen Yan, Kun Zhou, Lingjun Mao,
  Ruobing Han, Jinglin Cao, Wenpeng Xu, Lukun He, Minghao Fu, Fan Feng, and
  Biwei Huang.
\newblock {Video2World}: Benchmarking coding agents for interactive world
  modeling from embodied videos.
\newblock \emph{arXiv preprint arXiv:2610.04432}, 2026{\natexlab{a}}.

\bibitem[Tang et~al.(2026{\natexlab{b}})Tang, Shimada, Meng, Bi, Liu, Wang,
  Xiao, Tan, Zhang, Huang, Liang, Shen, Song, Vosoughi, Feng, Filvantorkaman,
  and Xu]{tang2026bvb}
Yolo~Y. Tang, Daiki Shimada, Jiayue Meng, Jing Bi, Pinxin Liu, Yicheng Wang,
  Yunzhong Xiao, Zhangyun Tan, Zeliang Zhang, Chao Huang, Susan Liang,
  Qianxiang Shen, Luchuan Song, Ali Vosoughi, Mingqian Feng, Melika
  Filvantorkaman, and Chenliang Xu.
\newblock {BVB}: Benchmarking agentic video understanding via programmatic
  reconstruction in blender.
\newblock \emph{arXiv preprint arXiv:2609.15478}, 2026{\natexlab{b}}.

\bibitem[Todorov et~al.(2012)Todorov, Erez, and Tassa]{todorov2012mujoco}
Emanuel Todorov, Tom Erez, and Yuval Tassa.
\newblock {MuJoCo}: A physics engine for model-based control.
\newblock In \emph{2012 IEEE/RSJ International Conference on Intelligent Robots
  and Systems}, pages 5026--5033, 2012.

\bibitem[Wang et~al.(2024)Wang, Xian, Chen, Wang, Wang, Fragkiadaki, Erickson,
  Held, and Gan]{wang2024robogen}
Yufei Wang, Zhou Xian, Feng Chen, Tsun-Hsuan Wang, Yian Wang, Katerina
  Fragkiadaki, Zackory Erickson, David Held, and Chuang Gan.
\newblock {RoboGen}: Towards unleashing infinite data for automated robot
  learning via generative simulation.
\newblock In \emph{International Conference on Machine Learning}, 2024.

\bibitem[Xia et~al.(2025)Xia, Su, Memmel, Jain, Yu, Mbiziwo-Tiapo, Farhadi,
  Gupta, Wang, and Ma]{xia2025drawer}
Hongchi Xia, Entong Su, Marius Memmel, Arhan Jain, Raymond Yu, Numfor
  Mbiziwo-Tiapo, Ali Farhadi, Abhishek Gupta, Shenlong Wang, and Wei-Chiu Ma.
\newblock {DRAWER}: Digital reconstruction and articulation with environment
  realism.
\newblock In \emph{Proceedings of the IEEE/CVF Conference on Computer Vision
  and Pattern Recognition}, pages 21771--21782, 2025.

\bibitem[Xiang et~al.(2020)Xiang, Qin, Mo, Xia, Zhu, Liu, Liu, Jiang, Yuan,
  Wang, Yi, Chang, Guibas, and Su]{xiang2020sapien}
Fanbo Xiang, Yuzhe Qin, Kaichun Mo, Yikuan Xia, Hao Zhu, Fangchen Liu, Minghua
  Liu, Hanxiao Jiang, Yifu Yuan, He Wang, Li Yi, Angel~X. Chang, Leonidas~J.
  Guibas, and Hao Su.
\newblock {SAPIEN}: A simulated part-based interactive environment.
\newblock In \emph{Proceedings of the IEEE/CVF Conference on Computer Vision
  and Pattern Recognition}, pages 11094--11104, 2020.

\bibitem[Yang et~al.(2024{\natexlab{a}})Yang, Jimenez, Wettig, Lieret, Yao,
  Narasimhan, and Press]{yang2024sweagent}
John Yang, Carlos~E. Jimenez, Alexander Wettig, Kilian Lieret, Shunyu Yao,
  Karthik Narasimhan, and Ofir Press.
\newblock {SWE-agent}: Agent-computer interfaces enable automated software
  engineering.
\newblock In \emph{Advances in Neural Information Processing Systems},
  2024{\natexlab{a}}.

\bibitem[Yang et~al.(2024{\natexlab{b}})Yang, Sun, Weihs, VanderBilt, Herrasti,
  Han, Wu, Haber, Krishna, Liu, Callison-Burch, Yatskar, Kembhavi, and
  Clark]{yang2024holodeck}
Yue Yang, Fan-Yun Sun, Luca Weihs, Eli VanderBilt, Alvaro Herrasti, Winson Han,
  Jiajun Wu, Nick Haber, Ranjay Krishna, Lingjie Liu, Chris Callison-Burch,
  Mark Yatskar, Aniruddha Kembhavi, and Christopher Clark.
\newblock {Holodeck}: Language guided generation of {3D} embodied {AI}
  environments.
\newblock In \emph{Proceedings of the IEEE/CVF Conference on Computer Vision
  and Pattern Recognition}, pages 16227--16237, 2024{\natexlab{b}}.

\bibitem[Yeshwanth et~al.(2023)Yeshwanth, Liu, Nie{\ss}ner, and
  Dai]{yeshwanth2023scannetpp}
Chandan Yeshwanth, Yueh-Cheng Liu, Matthias Nie{\ss}ner, and Angela Dai.
\newblock {ScanNet++}: A high-fidelity dataset of {3D} indoor scenes.
\newblock In \emph{Proceedings of the IEEE/CVF International Conference on
  Computer Vision}, pages 12--22, 2023.

\bibitem[Zhao et~al.(2026)Zhao, Zhou, Yang, Tang, Jian, Yao, Yao, Lin, Guo,
  Chen, Lyu, Ma, Wang, and Zhu]{zhao2026sceneactbench}
Yifei Zhao, Xiangxin Zhou, Wenhao Yang, Jiaqi Tang, Pu Jian, Huanjin Yao,
  Jiarui Yao, Haowei Lin, Chunchao Guo, Zhuo Chen, Wenkai Lyu, Jianzhu Ma,
  Xueqian Wang, and Wenxi Zhu.
\newblock {SceneActBench}: Can agents act on the {3D} scenes they see?
\newblock \emph{arXiv preprint arXiv:2607.22393}, 2026.

\bibitem[Zhou et~al.(2026)Zhou, Li, Lyu, Song, Huang, Zheng, Rupprecht,
  Vedaldi, and Wu]{zhou2026articraft}
Matt Zhou, Ruining Li, Xiaoyang Lyu, Zhaomou Song, Zhening Huang, Chuanxia
  Zheng, Christian Rupprecht, Andrea Vedaldi, and Shangzhe Wu.
\newblock {Articraft}: An agentic system for scalable articulated {3D} asset
  generation.
\newblock \emph{arXiv preprint arXiv:2605.15187}, 2026.

\end{thebibliography}
}

\clearpage
\setcounter{page}{1}
\renewcommand{\thepage}{S\arabic{page}}
\maketitlesupplementary
\appendix
\setcounter{table}{0}
\renewcommand{\thetable}{S\arabic{table}}
\renewcommand{\theHtable}{S\arabic{table}}
\setcounter{figure}{0}
\renewcommand{\thefigure}{S\arabic{figure}}
\renewcommand{\theHfigure}{S\arabic{figure}}
\arrayrulecolor{rifttabrule}
\setlength{\dbltextfloatsep}{10pt plus 1pt minus 1pt}
\section{Tasks and Reference Data}
\label{sec:supp-tasks}

\subsection{Task Specifications and Data Sources}
\label{sec:supp-task-specification}
\suppressfloats[t]
\begin{table}[t]
  \centering
  \caption{\textbf{Task composition and reference annotations across \projectname{}100.} Each category comprises 10 photographs evaluated across MuJoCo, Isaac Sim, and SAPIEN, reporting distinct capture sessions or rooms, scored articulated parts with revolute and prismatic splits, and scored oriented 3D furniture boxes.}
  \label{tab:task-inventory}
  \footnotesize
  \riftColSep{2pt}
  \renewcommand{\arraystretch}{1.15}
  \begin{tabularx}{\columnwidth}{@{}l*{4}{>{\raggedleft\arraybackslash}X}@{}}
    \toprule
    \textbf{Category} & \textbf{Scenes} & \textbf{Parts} & \textbf{Rev./Pris.} & \textbf{Boxes} \\
    \midrule
    \rowcolor{rifttabgroup}
    \multicolumn{5}{@{}l}{\textbf{\emph{Household30}}} \\
    Bedroom & 10 & 39 & 6\,/\,33 & 37 \\
    Living room & 10 & 47 & 2\,/\,45 & 48 \\
    Kitchen & 10 & 79 & 57\,/\,22 & 70 \\
    \cmidrule{1-5}
    \textbf{Subtotal} & \textbf{30} & \textbf{165} & \textbf{65\,/\,100} & \textbf{155} \\
    \midrule
    \rowcolor{rifttabgroup}
    \multicolumn{5}{@{}l}{\textbf{\emph{Workplace70}}} \\
    Conference room & 10 & 14 & 6\,/\,8 & 18 \\
    Classroom & 5 & 15 & 13\,/\,2 & 22 \\
    Office & 10 & 30 & 4\,/\,26 & 30 \\
    Laboratory & 10 & 36 & 13\,/\,23 & 23 \\
    Storage room & 7 & 34 & 9\,/\,25 & 18 \\
    Copy/mail room & 8 & 18 & 3\,/\,15 & 20 \\
    Lounge & 8 & 11 & 11\,/\,0 & 27 \\
    \cmidrule{1-5}
    \textbf{Subtotal} & \textbf{58} & \textbf{158} & \textbf{59\,/\,99} & \textbf{158} \\
    \midrule
    \rowcolor{rifttabgroup}
    \textbf{Total} & \textbf{88} & \textbf{323} & \textbf{124\,/\,199} & \textbf{313} \\
    \bottomrule
  \end{tabularx}
\end{table}

\paragraph{Data sources and preprocessing.}
Table~\ref{tab:task-inventory} details the 30 residential tasks from distinct ARKitScenes~\cite{baruch2021arkitscenes} sessions and 70 workplace tasks across 58 rooms from ScanNet++ v2~\cite{yeshwanth2023scannetpp} and MuSHRoom~\cite{ren2024mushroom}. The lounge category contains six ScanNet++ photographs from six rooms and four MuSHRoom photographs from two rooms, while all other workplace categories use ScanNet++ v2. Same-room ScanNet++ frames remain at least $1\,\text{m}$ or $45^\circ$ apart, and the articulation term is omitted on the 23 workplace tasks without scored movable parts. Test-time inputs consist solely of single EXIF-stripped RGB images kept at $1920\times1440$ for ARKitScenes, undistorted and centrally cropped to $4{:}3$ without rescaling for ScanNet++, and losslessly cropped to $736\times992$ for MuSHRoom.

\paragraph{Submission contract and coordinate conventions.}
Submissions pair native scene descriptions in MuJoCo MJCF~\cite{todorov2012mujoco}, Isaac Sim OpenUSD~\cite{nvidia2025isaacsim}, or SAPIEN assembly format~\cite{xiang2020sapien} with deterministic Python build scripts, procedural geometry and materials, and structured camera, object, and joint declarations. All 100 tasks share a right-handed metric frame with the floor at the origin beneath the estimated camera, $+Z$ upward, $+Y$ forward, and $+X$ rightward. Articulated furniture must be authored purely in code without external 3D assets or pretrained vision models, incorporating hollow cavities with at least $8\,\text{cm}$ door clearance, movable drawer boxes, and collidable handles.

\subsection{Annotation Details and Data Release}
\label{sec:supp-references}

\paragraph{Filtering thresholds and auxiliary masks.}
Oriented 3D boxes for fixed furniture and appliances are refitted within $2\,\text{cm}$ of laser scans while excluding soft goods, retaining boxes above $1.5\%$ image area for scoring. Category conventions standardize ambiguous structural boundaries, whereas auxiliary masks isolate obstructed openings, open shelving, and glass or specular surfaces lacking valid scan depth so that no metric reads missing returns. From the locked ground truth, each task also instantiates \textbf{E{\scriptsize MPTY}}, \textbf{S{\scriptsize HELL}}, and \textbf{O{\scriptsize RACLE}} reference controls validated in Sec.~\ref{sec:supp-quality}.

\paragraph{Licensing and data release.}
Source datasets follow non-commercial terms for ARKitScenes, CC BY 4.0 for SceneFun3D, Articulate3D, and MuSHRoom, and the ScanNet++ Terms of Use. The benchmark codebase, frozen evaluator, and reference annotations are publicly available on \href{\riftCodeURL}{GitHub} and \href{\riftDataURL}{Hugging Face} under licenses compatible with the source datasets, with ScanNet++ images and scan-derived references regenerated locally from the official ScanNet++ release via provided scripts.

\section{Evaluation Details}
\label{sec:supp-evaluation}

\subsection{Execution Environment and Public Feedback}
\label{sec:supp-execution-protocol}

\begin{figure*}[!t]
  \centering
  \footnotesize
  \setlength{\tabcolsep}{8pt}
  \renewcommand{\arraystretch}{1.15}
  \begin{tabular}{p{\dimexpr\textwidth-2\tabcolsep\relax}}
    \toprule
    \rowcolor{rifttabgroup}
    \rule{0pt}{2.7ex}{\fontsize{11}{13}\selectfont\bfseries Standardized Single-Image Interactive Reconstruction Prompt}\rule[-0.8ex]{0pt}{0pt} \\
    \midrule
    \textbf{Task Goal} \\[2pt]
    Given one EXIF-stripped indoor photograph, write code from scratch to construct a simulation scene that \emph{runs} stably under gravity, \emph{can be operated} by a robot arm grasping handles to open doors and drawers via physical joints, and \emph{matches the photograph} in layout, geometry, and appearance from the estimated camera. \\
    \addlinespace[6pt]
    \textbf{Workspace \& Hard Rules} \\[2pt]
    Operate offline inside an isolated workspace equipped with numerical and mesh libraries, a public self-check suite, and the reference Franka Panda arm. Pretrained vision or 3D models, external 3D asset repositories, texture projection beyond procedural materials or local crops up to $128\times128$\,px, disabling gravity or contacts, and special-casing the checker are prohibited. Full-resolution image crops may be inspected in code to resolve fine door gaps and handles. \\
    \addlinespace[6pt]
    \textbf{Frame \& Physics} \\[2pt]
    Place the floor at $z=0$ beneath the estimated camera in a right-handed metric frame with $+Z$ up, $+Y$ forward, and $+X$ right. Model room surfaces, furniture, and appliances as fixed bodies and loose tabletop items as 6-DoF free bodies with realistic masses. Build hollow carcasses and drawers from multiple convex panels with $2\text{--}3\,\text{mm}$ clearances rather than solid convex hulls, set joint state $0$ to the closed photograph pose with positive opening travel, and attach collidable handles. \\
    \addlinespace[6pt]
    \textbf{Scoring \& Workflow} \\[2pt]
    Overall quality combines articulated joints, object geometry, spatial layout, and unlit appearance, scaled multiplicatively by physical validity, interior completeness, and global scale accuracy, with robot manipulation reported separately. Deliver a loadable submission early and refine it iteratively via full-resolution visual overlays and physics self-checks. \\
    \bottomrule
  \end{tabular}
\end{figure*}

Coding agents execute inside version-pinned headless containers provisioned solely with the input photograph and the standardized prompt summarized above, while hidden ground-truth scans, annotations, and the evaluator remain unmounted and outbound traffic is restricted to the model inference endpoint. Although modeling conventions and metric weights are disclosed transparently, articulation credit requires accurate 3D camera and layout registration and couples multiplicatively with physical stability and interior completeness, precluding test-time score probing or isolated sub-metric optimization.

\paragraph{Container environment and rendering stack.}
Containers preinstall scientific and mesh-processing packages alongside headless offscreen renderers using EGL in MuJoCo, RTX or Vulkan in Isaac Sim, and Vulkan in SAPIEN, stepping at $\le 5\,\text{ms}$ under standard gravity. MuJoCo and SAPIEN containers allocate two CPU cores, 8\,GiB of RAM, and 10\,GiB of storage, whereas Isaac Sim allocates four CPU cores and 16\,GiB of RAM with pre-clock warmup. Shell commands carry a 2-minute default timeout extendable to 10 minutes for full robot self-checks, with unconstrained tool-call counts within the 240-minute active budget.

\paragraph{Modeling constraints and public self-checks.}
Free-body and articulated-part centers of mass must reside within $\max(2\,\text{mm},2\%)$ of their collider diagonals with valid positive inertias and massless decorative geometry. Gravity compensation, tendons, equality constraints, hidden actuators, sleep states, opening springs, negative damping, and contact exclusions outside a movable part's own static carcass are prohibited. Without accessing ground truth, the public self-check suite provides full-resolution side-by-side and overlay renders from the estimated camera, a $3\,\text{s}$ gravitational settling test, full-range joint sweeps paired with a $20\,\text{N}$ pull over $3\,\text{s}$, ray checks for $\ge 8\,\text{cm}$ cabinet clearance and drawer bottoms, and simulated Franka Panda grasp-and-open rollouts.

\subsection{Submission, Reproduction, and Cost Accounting}
\label{sec:supp-submissions}

\paragraph{Active-time metering and clean rebuild.}
Each task enforces a 240-minute budget of \emph{Active Elapsed Time} covering inference, ordinary latency, rate-limit retries, and command execution while deducting quota pauses, verified outages, and 15-minute checkpoint exports. Offline clean-rebuild verification copies each submission into a network-isolated read-only container, removes all pre-generated scenes, meshes, textures, and checker outputs, and reruns the build script from scratch under a fixed timeout, assigning $S_{\text{rebuilt}}=0$ upon absolute paths, symbolic links, runtime errors, missing outputs, or timeouts while isolating host faults.

\paragraph{Resource and code accounting.}
The harness and API proxy log input, output, and cached tokens alongside tool invocations, converting usage into monetary cost under official provider rates and counting non-blank, non-comment source lines in the final build scripts.

\subsection{Scoring and Interaction Tests}
\label{sec:supp-quality}

\begin{figure*}[t]
  \centering
  \includegraphics[pagebox=cropbox,width=\textwidth]{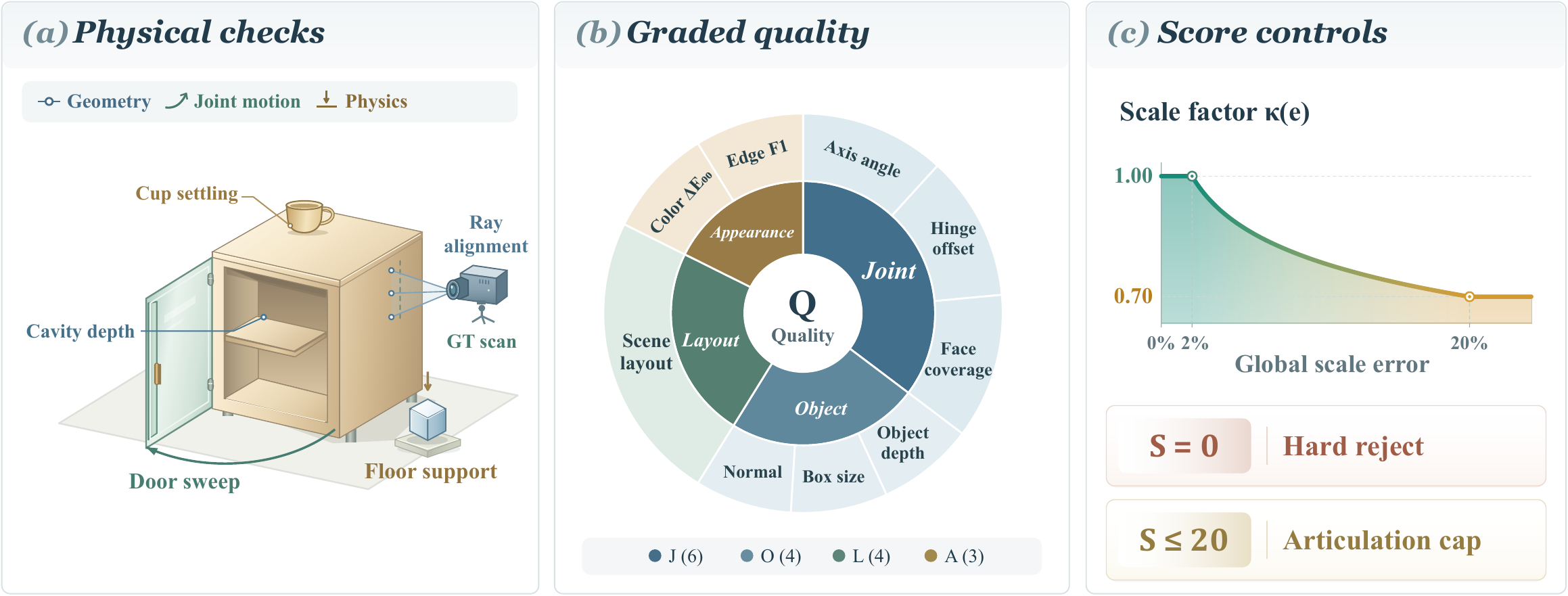}
  \caption{\textbf{Overview of the \projectname{} evaluation pipeline.} Panels~a--c summarize geometric and physical checks, weighted quality components with $\mathbf{w}=(6,4,4,3)$, and multiplicative scale and failure-cap controls.}
  \label{fig:supp-scoring-overview}
\end{figure*}

\paragraph{Sub-metric design and interaction diagnostics.}
As illustrated in Figure~\ref{fig:supp-scoring-overview}, geometric and physical metrics evaluate collision geometry rendered from the submitted camera, whereas visual-only geometry contributes solely to appearance $A$. Beyond the quality components summarized in Figure~\ref{fig:supp-scoring-overview}b, articulation $J$ scales part credit by matched-joint precision to penalize spurious joints, object geometry $O$ penalizes floating false positives and weights annotated small items equally, spatial layout $L$ measures depth agreement relative to a zero-credit median-depth plane alongside floor IoU, and appearance $A$ normalizes exposure prior to comparing unlit color and edge structure while zeroing whole-photo projection. Diagnostic rates in Figure~\ref{fig:failure-diagnostics-heatmap} use applicable denominators, and separate Franka Panda tests approach declared handles within a $3\,\text{cm}$ recessed-pull tolerance across all 323 scored parts.

\paragraph{Scoring fairness and cross-engine parity.}
All three simulators enforce identical ray-sampling density, interpenetration thresholds, and torque schedules. Across 100 MuJoCo tasks, mean reference scores rise monotonically from 7.0 on \textbf{E{\scriptsize MPTY}} to 47.2 on \textbf{S{\scriptsize HELL}} and 87.9 on \textbf{O{\scriptsize RACLE}}, strictly increasing on 94 tasks and tying on six scenes without hidden completions. Joint-free \textbf{E{\scriptsize MPTY}} shells remain capped within 20 points, cavity-free \textbf{S{\scriptsize HELL}} scenes are halved by $C=0$ on 89 tasks, and \textbf{O{\scriptsize RACLE}} exceeds 0.95 on geometry, articulation, and physics while falling short of 100 solely due to untextured solid-color primitives. Across MuJoCo, Isaac Sim, and SAPIEN, both 100-task shared-representation baselines such as SAM 3D Objects at 10.56, 10.56, and 10.44 and converted \emph{Ablation30} controls at 7.4, 7.4, and 7.6 on \textbf{E{\scriptsize MPTY}}, 48.5, 48.5, and 48.1 on \textbf{S{\scriptsize HELL}}, and 88.0, 88.0, and 87.2 on \textbf{O{\scriptsize RACLE}} match closely with identical per-task ordering, confirming that simulator gaps reflect interface friction rather than evaluator bias.

\section{Additional Experiments}
\label{sec:supp-experiments}

\begin{table*}[t]
  \centering
  \caption{\textbf{Per-simulator overall and component scores on \projectname{}100.} Sub-metrics follow Table~\ref{tab:metric-definitions} on a 0--100 scale and are averaged over all 100 tasks, counting failed tasks and tasks without the scored component as zero, with per-block best entries bolded and slashes denoting non-agent baselines.}
  \label{tab:component-results}
  \footnotesize
  \riftColSep{2pt}
  \renewcommand{\arraystretch}{1.16}
  \begin{tabular*}{\textwidth}{@{\extracolsep{\fill}}ll*{7}{r}@{}}
    \toprule
    \textbf{Model / Method} & \textbf{Harness} & \textbf{Overall}\,\uparr & \boldmath$O$\,\uparr & \boldmath$L$\,\uparr & \boldmath$A$\,\uparr & \boldmath$J$\,\uparr & \boldmath$P$\,\uparr & \boldmath$C$\,\uparr \\
    \midrule
    \rowcolor{rifttabgroup}
    \multicolumn{9}{@{}l}{\textbf{MuJoCo}} \\
    Claude Opus 5.5 & Claude Code & \textbf{56.62} & \textbf{55.8} & \textbf{42.3} & \textbf{84.8} & \textbf{54.6} & \textbf{100.0} & \textbf{95.4} \\
    GPT-6 Astra & Codex & 39.52 & 41.9 & 30.2 & 79.6 & 32.8 & 99.9 & 89.1 \\
    Kimi K3.1 Pro & Kimi Code & 26.29 & 25.2 & 17.2 & 74.1 & 24.9 & 97.7 & 84.9 \\
    DeepSeek-V4.1-Flash-1017 & DSH & 22.36 & 17.9 & 15.6 & 65.4 & 20.1 & 92.6 & 73.2 \\
    GLM 5.3 Flash & ZCode & 22.55 & 18.6 & 15.5 & 68.6 & 20.5 & 96.7 & 80.7 \\
    MiniMax M3.1-Flash & MiniMax Code & 18.76 & 15.5 & 13.4 & 60.7 & 19.6 & 89.9 & 68.5 \\
    SAM 3D Objects & / & 10.56 & 19.2 & 12.0 & 59.2 & 0.0 & 64.4 & 66.1 \\
    SAM 3D Objects + Particulate & / & 6.76 & 19.0 & 12.2 & 62.3 & 0.4 & 45.2 & 63.3 \\
    MIDI & / & 4.51 & 5.5 & 6.0 & 48.0 & 0.0 & 48.1 & 31.5 \\
    URDFormer & / & 2.60 & 6.2 & 4.9 & 26.7 & 1.0 & 42.7 & 20.8 \\
    \midrule
    \rowcolor{rifttabgroup}
    \multicolumn{9}{@{}l}{\textbf{Isaac Sim}} \\
    Claude Opus 5.5 & Claude Code & \textbf{51.10} & \textbf{50.2} & \textbf{38.0} & 81.1 & \textbf{48.1} & 99.0 & \textbf{96.5} \\
    GPT-6 Astra & Codex & 41.57 & 43.8 & 30.0 & \textbf{84.3} & 38.2 & \textbf{100.0} & 95.1 \\
    Kimi K3.1 Pro & Kimi Code & 23.55 & 21.0 & 15.0 & 66.5 & 21.5 & 96.0 & 81.1 \\
    DeepSeek-V4.1-Flash-1017 & DSH & 20.21 & 15.5 & 15.1 & 55.7 & 20.1 & 91.0 & 72.9 \\
    GLM 5.3 Flash & ZCode & 16.81 & 16.0 & 11.2 & 51.2 & 16.5 & 91.4 & 64.4 \\
    MiniMax M3.1-Flash & MiniMax Code & 11.43 & 11.4 & 9.4 & 37.5 & 12.1 & 68.8 & 50.4 \\
    SAM 3D Objects & / & 10.56 & 19.2 & 12.1 & 59.2 & 0.0 & 64.4 & 66.1 \\
    SAM 3D Objects + Particulate & / & 6.63 & 19.0 & 12.2 & 62.3 & 0.5 & 43.8 & 63.3 \\
    MIDI & / & 4.49 & 5.5 & 6.1 & 48.0 & 0.0 & 48.1 & 31.5 \\
    URDFormer & / & 2.61 & 6.2 & 4.9 & 26.7 & 1.1 & 42.9 & 21.7 \\
    \midrule
    \rowcolor{rifttabgroup}
    \multicolumn{9}{@{}l}{\textbf{SAPIEN}} \\
    Claude Opus 5.5 & Claude Code & \textbf{50.72} & \textbf{52.5} & \textbf{37.2} & \textbf{81.8} & \textbf{48.0} & \textbf{100.0} & \textbf{95.0} \\
    GPT-6 Astra & Codex & 37.62 & 40.8 & 26.6 & 80.1 & 32.3 & 98.5 & 86.3 \\
    Kimi K3.1 Pro & Kimi Code & 26.01 & 25.1 & 18.0 & 72.1 & 27.5 & 99.6 & 85.6 \\
    DeepSeek-V4.1-Flash-1017 & DSH & 22.31 & 19.3 & 17.4 & 64.5 & 17.6 & 96.0 & 81.1 \\
    GLM 5.3 Flash & ZCode & 17.23 & 13.8 & 13.5 & 59.4 & 16.0 & 99.4 & 75.3 \\
    MiniMax M3.1-Flash & MiniMax Code & 16.37 & 14.7 & 12.3 & 55.3 & 14.7 & 92.0 & 70.4 \\
    SAM 3D Objects & / & 10.44 & 19.4 & 12.0 & 59.3 & 1.0 & 64.9 & 67.1 \\
    SAM 3D Objects + Particulate & / & 6.75 & 19.2 & 12.2 & 62.5 & 3.5 & 45.7 & 62.4 \\
    MIDI & / & 4.53 & 5.5 & 6.4 & 49.3 & 0.0 & 49.6 & 32.5 \\
    URDFormer & / & 2.51 & 6.2 & 4.9 & 26.3 & 1.0 & 42.7 & 19.2 \\
    \bottomrule
  \end{tabular*}
\end{table*}

\begin{table*}[t]
  \centering
  \caption{\textbf{Per-simulator failure diagnostics on \projectname{}100.} Entries report percentage failure rates following Figure~\ref{fig:failure-diagnostics-heatmap}, with dashes denoting metrics inapplicable to static or non-agent baselines.}
  \label{tab:supp-failure-diagnostics}
  \footnotesize
  \riftColSep{1.2pt}
  \renewcommand{\arraystretch}{1.16}
  \begin{tabular*}{\textwidth}{@{\extracolsep{\fill}}lccc@{\hspace{6pt}}cccc@{\hspace{6pt}}cc@{}}
    \toprule
    & & \multicolumn{2}{c@{\hspace{6pt}}}{\textbf{Object \& Layout}}
    & \multicolumn{4}{c@{\hspace{6pt}}}{\textbf{Articulation \& Physics}}
    & \multicolumn{2}{c}{\textbf{Reproducibility}} \\
    \cmidrule(lr{7pt}){3-4} \cmidrule(lr{7pt}){5-8} \cmidrule(l){9-10}
    \textbf{Model / Method} & \textbf{Harness}
    & \shortstack{{\scriptsize Object}\\{\scriptsize Omission\,\dnarr}}
    & \shortstack{{\scriptsize Position}\\{\scriptsize Error\,\dnarr}}
    & \shortstack{{\scriptsize Part}\\{\scriptsize Omission\,\dnarr}}
    & \shortstack{{\scriptsize Joint}\\{\scriptsize Error\,\dnarr}}
    & \shortstack{{\scriptsize Motion}\\{\scriptsize Failure\,\dnarr}}
    & \shortstack{{\scriptsize Rest}\\{\scriptsize Penetration\,\dnarr}}
    & \shortstack{{\scriptsize Invalid}\\{\scriptsize Submission\,\dnarr}}
    & \shortstack{{\scriptsize Rebuild}\\{\scriptsize Failure\,\dnarr}} \\
    \midrule
    \rowcolor{rifttabgroup}
    \multicolumn{10}{@{}l}{\textbf{MuJoCo}} \\
    Claude Opus 5.5 & Claude Code & 10.5 & 70.4 & 5.3 & 68.3 & 3.9 & 0.0 & 0.0 & 0.0 \\
    GPT-6 Astra & Codex & 25.6 & 79.0 & 18.9 & 84.0 & 8.8 & 0.0 & 0.0 & 0.0 \\
    Kimi K3.1 Pro & Kimi Code & 24.3 & 88.3 & 24.3 & 95.4 & 21.5 & 0.0 & 0.0 & 0.0 \\
    DeepSeek-V4.1-Flash-1017 & DSH & 36.6 & 91.3 & 31.6 & 94.5 & 31.5 & 0.0 & 1.0 & 2.0 \\
    GLM 5.3 Flash & ZCode & 35.4 & 90.8 & 33.8 & 94.6 & 37.6 & 0.0 & 0.0 & 0.0 \\
    MiniMax M3.1-Flash & MiniMax Code & 40.0 & 94.4 & 38.7 & 94.6 & 18.9 & 5.4 & 5.0 & 5.0 \\
    SAM 3D Objects & / & 44.9 & 95.2 & 100.0 & -- & -- & 0.0 & 4.0 & -- \\
    SAM 3D Objects + Particulate & / & 44.9 & 93.6 & 79.3 & 100.0 & 91.0 & 84.0 & 0.0 & -- \\
    MIDI & / & 75.3 & 98.6 & 100.0 & -- & -- & 0.0 & 3.0 & -- \\
    URDFormer & / & 84.3 & 87.8 & 78.9 & 100.0 & 83.8 & 11.0 & 0.0 & -- \\
    \midrule
    \rowcolor{rifttabgroup}
    \multicolumn{10}{@{}l}{\textbf{Isaac Sim}} \\
    Claude Opus 5.5 & Claude Code & 12.5 & 78.3 & 4.6 & 73.4 & 19.5 & 0.0 & 0.0 & 0.0 \\
    GPT-6 Astra & Codex & 11.8 & 82.6 & 11.5 & 87.1 & 11.9 & 1.0 & 0.0 & 0.0 \\
    Kimi K3.1 Pro & Kimi Code & 33.9 & 94.0 & 29.6 & 98.2 & 33.6 & 1.0 & 0.0 & 0.0 \\
    DeepSeek-V4.1-Flash-1017 & DSH & 43.3 & 92.3 & 30.6 & 95.7 & 45.5 & 0.0 & 1.0 & 6.0 \\
    GLM 5.3 Flash & ZCode & 46.9 & 94.9 & 40.1 & 96.6 & 36.5 & 6.1 & 0.0 & 0.0 \\
    MiniMax M3.1-Flash & MiniMax Code & 47.9 & 88.0 & 46.4 & 97.9 & 33.1 & 16.9 & 14.0 & 15.0 \\
    SAM 3D Objects & / & 44.9 & 95.2 & 100.0 & -- & -- & 0.0 & 4.0 & -- \\
    SAM 3D Objects + Particulate & / & 44.9 & 93.6 & 79.3 & 100.0 & 88.1 & 83.5 & 0.0 & -- \\
    MIDI & / & 75.4 & 98.6 & 100.0 & -- & -- & 0.0 & 3.0 & -- \\
    URDFormer & / & 84.3 & 87.8 & 78.9 & 100.0 & 82.4 & 11.0 & 0.0 & -- \\
    \midrule
    \rowcolor{rifttabgroup}
    \multicolumn{10}{@{}l}{\textbf{SAPIEN}} \\
    Claude Opus 5.5 & Claude Code & 10.2 & 77.6 & 3.7 & 84.9 & 5.1 & 0.0 & 0.0 & 0.0 \\
    GPT-6 Astra & Codex & 22.7 & 81.0 & 22.0 & 92.1 & 11.9 & 0.0 & 0.0 & 0.0 \\
    Kimi K3.1 Pro & Kimi Code & 24.9 & 92.3 & 25.4 & 96.7 & 34.9 & 1.0 & 0.0 & 0.0 \\
    DeepSeek-V4.1-Flash-1017 & DSH & 35.4 & 93.5 & 37.2 & 96.9 & 35.7 & 0.0 & 1.0 & 4.0 \\
    GLM 5.3 Flash & ZCode & 43.5 & 94.9 & 40.6 & 97.9 & 51.0 & 1.0 & 0.0 & 0.0 \\
    MiniMax M3.1-Flash & MiniMax Code & 44.5 & 93.4 & 38.7 & 98.4 & 39.6 & 7.4 & 6.0 & 6.0 \\
    SAM 3D Objects & / & 44.7 & 95.2 & 100.0 & -- & -- & 0.0 & 4.0 & -- \\
    SAM 3D Objects + Particulate & / & 44.7 & 93.6 & 79.6 & 100.0 & 84.8 & 84.0 & 0.0 & -- \\
    MIDI & / & 75.8 & 98.6 & 100.0 & -- & -- & 0.0 & 3.0 & -- \\
    URDFormer & / & 84.3 & 87.8 & 78.9 & 100.0 & 83.8 & 11.0 & 0.0 & -- \\
    \bottomrule
  \end{tabular*}
\end{table*}

\begin{table*}[!t]
  \centering
  \caption{\textbf{Coding agent runtime configurations.} Native evaluations use official CLI defaults under \emph{High} reasoning effort, while controlled ablations standardize token and compaction budgets across models.}
  \label{tab:supp-agent-configs}
  \footnotesize
  \riftColSep{0pt}
  \renewcommand{\arraystretch}{1.16}
  \begin{tabular*}{\textwidth}{@{\extracolsep{\fill}}llcccc@{}}
    \toprule
    \textbf{Model} & \textbf{CLI Harness} & \textbf{Effort} & \textbf{Context} & \textbf{Compaction} & \textbf{Max Output} \\
    \midrule
    \rowcolor{rifttabgroup}\multicolumn{6}{@{}l}{\textbf{\emph{Native Harness Configurations}}} \\
    Claude Opus 5.5 & Claude Code v2.1.314 & High & Official default & Official default & Official default \\
    GPT-6 Astra & Codex v0.165.0 & High & Official default & Official default & Official default \\
    Kimi K3.1 Pro & Kimi Code v2.4.0 & High & Official default & Official default & Official default \\
    DeepSeek-V4.1-Flash-1017 & DSH v0.2.1-rc.2 & High & Official default & Official default & Official default \\
    GLM 5.3 Flash & ZCode v0.16.9 & High & Official default & Official default & Official default \\
    MiniMax M3.1-Flash & MiniMax Code v0.6.5 & High & Official default & Official default & Official default \\
    \midrule
    \rowcolor{rifttabgroup}\multicolumn{6}{@{}l}{\textbf{\emph{Controlled Harness Ablations}}} \\
    \riftGroupFirst{2}{GPT-6 Astra, DeepSeek-V4.1-Flash-1017, GLM 5.3 Flash} & Claude Code v2.1.314 & High & 196,608 & Auto-compact & 65,536 \\
    \riftGroupLast{2}{\riftTwoLine{GPT-6 Astra, DeepSeek-V4.1-}{Flash-1017, GLM 5.3 Flash}} & \emph{mini-SWE-agent} v2.4.6 & High & Model limit & Disabled & 128,000 \\
    \bottomrule
  \end{tabular*}
\end{table*}
 
\begin{table}[t]
  \centering
  \caption{\textbf{Per-simulator reasoning-effort ablation on \emph{Ablation30}.} Mean \emph{Code-Rebuilt Score} $S_{\text{rebuilt}}$ under each native CLI harness across supported effort levels, with default \emph{High} settings shaded.}
  \label{tab:supp-effort-by-sim}
  \footnotesize
  \riftColSep{1pt}
  \renewcommand{\arraystretch}{1.16}
  \begin{tabularx}{\columnwidth}{@{}l<{\hspace{4pt}}l<{\hspace{2pt}}>{\hsize=1.08\hsize\linewidth=\hsize\centering\arraybackslash}X>{\hsize=0.94\hsize\linewidth=\hsize\centering\arraybackslash}X>{\hsize=1.1\hsize\linewidth=\hsize\centering\arraybackslash}X>{\hsize=0.88\hsize\linewidth=\hsize\centering\arraybackslash}X@{}}
    \toprule
    \textbf{Model} & \textbf{Effort} & {\fontsize{6.5pt}{7.5pt}\selectfont\textbf{MuJoCo}\uparr} & {\fontsize{6.5pt}{7.5pt}\selectfont\textbf{Isaac}\uparr} & {\fontsize{6.5pt}{7.5pt}\selectfont\textbf{SAPIEN}\uparr} & {\fontsize{6.5pt}{7.5pt}\selectfont\textbf{Mean}\uparr} \\
    \midrule
    \riftGroupFirst{6}{GPT-6 Astra} & Low & 35.42 & 37.25 & 38.91 & 37.19 \\
     & Medium & 34.57 & 40.11 & 34.50 & 36.39 \\
    \rowcolor{rifttabhi}
     & High & 39.44 & 41.54 & 38.14 & 39.71 \\
     & XHigh & 44.51 & \textbf{48.75} & 46.11 & 46.46 \\
     & Max & \textbf{50.51} & 48.31 & 44.51 & 47.78 \\
    \riftGroupLast{6}{GPT-6 Astra} & Ultra & 48.87 & 48.41 & \textbf{49.43} & \textbf{48.90} \\
    \midrule
    \riftGroupFirst{6}{\riftTwoLine{DeepSeek-V4.1-}{Flash-1017}} & Minimal & 22.99 & 20.96 & 20.93 & 21.63 \\
     & Low & 25.04 & 20.99 & 23.27 & 23.10 \\
     & Medium & 22.28 & 22.07 & 22.13 & 22.16 \\
    \rowcolor{rifttabhi}
     & High & 20.01 & 23.80 & \textbf{24.02} & 22.61 \\
     & XHigh & 21.75 & 22.49 & 22.72 & 22.32 \\
    \riftGroupLast{6}{\riftTwoLine{DeepSeek-V4.1-}{Flash-1017}} & Max & \textbf{26.77} & \textbf{25.59} & 22.38 & \textbf{24.91} \\
    \midrule
    \riftGroupFirst{3}{GLM 5.3 Flash} & Low & 14.48 & 11.00 & 11.74 & 12.41 \\
    \rowcolor{rifttabhi}
     & High & \textbf{23.43} & \textbf{17.59} & \textbf{19.58} & \textbf{20.20} \\
    \riftGroupLast{3}{GLM 5.3 Flash} & Max & 13.98 & 6.98 & 17.95 & 12.97 \\
    \bottomrule
  \end{tabularx}
\end{table}

\subsection{Experimental Setup}
\label{sec:supp-methods}

\paragraph{Specialized 3D baseline pipelines.}
All four open-source specialized baselines take only the single input photograph and share a frozen two-stage perception front end in which Grounding DINO~\emph{base} and SAM~2.1~\emph{hiera-large} extract up to 40 indoor object masks at threshold $0.30$ alongside a floor mask at $0.25$, while MoGe-2~\emph{vitl-normal} predicts a metric pointmap and camera intrinsics for RANSAC ground-plane fitting within $2\,\text{cm}$ and $30^\circ$. Each baseline then runs offline under official weights, where SAM 3D Objects reconstructs rigid meshes, Particulate adds part segmentation and joint prediction, and URDFormer and MIDI align outputs to the shared depth map, followed by CoACD~1.0.14 convex decomposition at seed $0$ and deterministic export into MuJoCo, Isaac Sim, and SAPIEN.

\paragraph{Coding agent runtime configurations.}
Table~\ref{tab:supp-agent-configs} summarizes runtime configurations across all six native setups and both controlled harness ablations. Native evaluations preserve each CLI's official default context window, compaction policy, and output limit at \emph{High} reasoning effort. For controlled ablations on GPT-6 Astra, GLM 5.3 Flash, and DeepSeek-V4.1-Flash-1017, the Claude Code setup enforces a shared 196{,}608-token context window with built-in auto-compaction and a 65{,}536-token output cap, while the \emph{mini-SWE-agent}~\cite{minisweagent2025} setup retains uncompacted linear history up to each model limit with a 128{,}000-token output cap. Every run executes in headless non-interactive mode under default sampling parameters with web access and external plugins disabled, enforcing per-attempt log caps of $8\,\text{MiB}$ per line, $256\,\text{MiB}$ cumulative output, and $512\,\text{MiB}$ session state. All commercial models are queried directly through official provider API endpoints. We check the scores and agent trajectories of every run, and runs affected by provider-side serving anomalies are re-run under the identical configuration.

\subsection{Detailed Quantitative Results}
\label{sec:supp-results}

\begin{figure}[!t]
  \centering
  \includegraphics[width=\columnwidth]{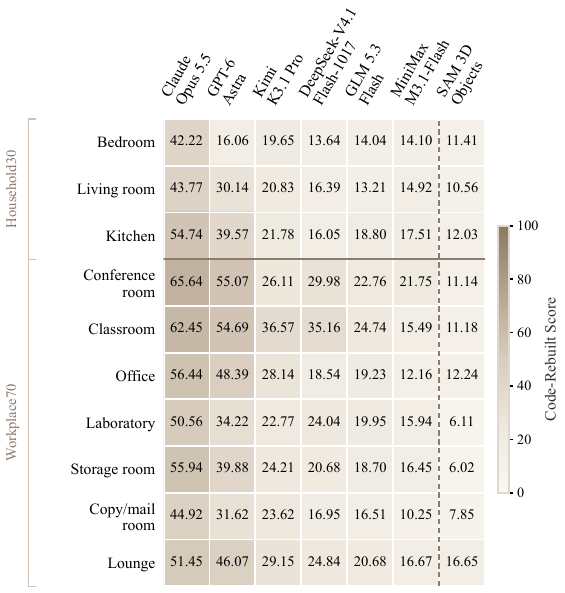}
  \caption{\textbf{Category-wise reconstruction scores on \projectname{}100.} Cells report three-simulator mean $S_{\text{rebuilt}}$ across the ten indoor categories, separated by residential versus workplace splits and coding agents versus SAM 3D Objects.}
  \label{fig:supp-category-results}
\end{figure}

\paragraph{Component and category breakdowns.}
Table~\ref{tab:component-results} and Table~\ref{tab:supp-failure-diagnostics} decompose three-simulator scores into quality sub-metrics, physical modifiers, and eight failure diagnostic rates, showing that closed-loop coding agents sustain physical validity $P \ge 68.8$ and completeness $C \ge 50.4$, reaching at least $91.0$ and $64.4$ for five of six models, whereas feed-forward baselines lose heavily on articulation and interior cavities. Across the ten indoor categories in Figure~\ref{fig:supp-category-results}, coding agents peak on modular workspaces such as \emph{Conference room} at 65.64 for Claude Opus 5.5 and \emph{Classroom} at 62.45, drop on cluttered \emph{Bedroom} at 42.22 and \emph{Living room} at 43.77, and surpass SAM 3D Objects by over ninefold on cabinet-rich \emph{Storage room} at 55.94 versus 6.02.

\paragraph{Per-simulator reasoning-effort results.}
Table~\ref{tab:supp-effort-by-sim} expands the three-simulator means of Table~\ref{tab:effort-ablation} into MuJoCo, Isaac Sim, and SAPIEN scores, showing that higher reasoning effort lifts GPT-6 Astra consistently across all three engines from 38.14 to 41.54 at \emph{High} into the 44.51 to 50.51 range at \emph{XHigh} through \emph{Ultra}, whereas GLM 5.3 Flash drops sharpest on Isaac Sim from 17.59 at \emph{High} to 6.98 at \emph{Max}.

\paragraph{Recurring failure archetypes.}
Across quantitative diagnostics and visual inspection, remaining bottlenecks concentrate into four recurring failure modes. First, monocular depth and placement drift leaves scenes matching the 2D camera silhouette while misplacing cabinets by tens of centimeters in 3D, driving $75.4\%\text{--}98.6\%$ object position error across evaluated methods. Second, symmetric double doors and flush handleless panels frequently trigger hinge-side or opening-axis inversion, accounting for $75.5\%\text{--}100.0\%$ joint-structure error. Third, dynamic collider self-locking causes scenes with near-zero static rest penetration to jam immediately under $20\,\text{N}$ pull actuation due to sub-centimeter door-frame overlap or unsplit convex-hull carcasses. Fourth, painted or overly thin pulls induce handle-contact slippage during simulated Franka Panda parallel-jaw grasping despite functional underlying joints.

\subsection{Efficiency and Time Budgets}
\label{sec:supp-comparisons}

\begin{figure}[t]
  \centering
  \includegraphics[width=\columnwidth]{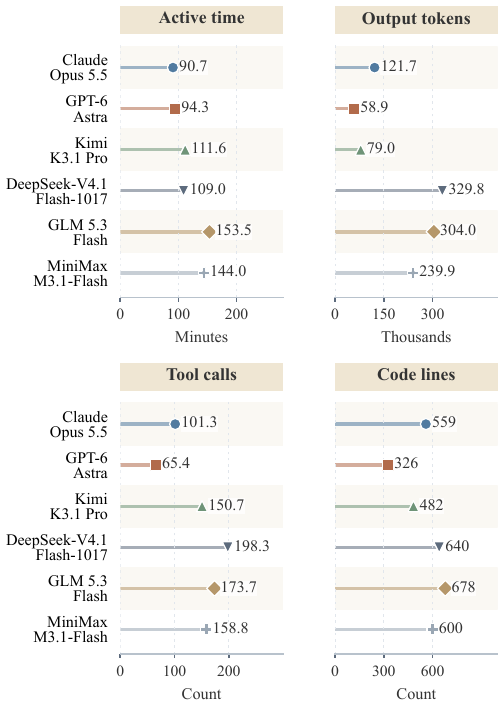}
  \caption{\textbf{Per-task resource usage across coding agents.} Panels report three-simulator mean active time, output tokens, tool calls, and source code lines on \projectname{}100 under each native harness.}
  \label{fig:supp-efficiency}
\end{figure}

\begin{figure}[t]
  \centering
  \includegraphics[width=\columnwidth]{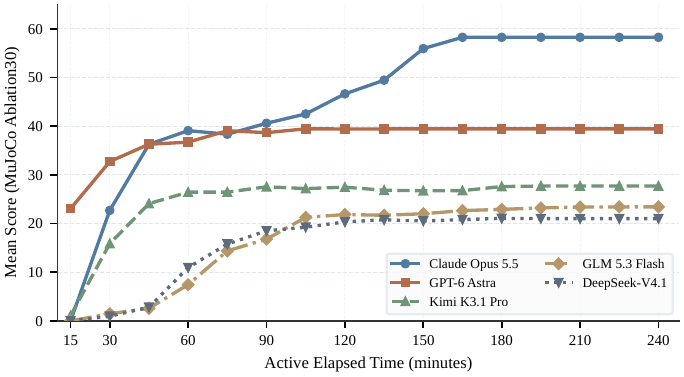}
  \caption{\textbf{Quality-vs-time trajectories from 15-minute workspace checkpoints on MuJoCo \emph{Ablation30}.} Early checkpoints without a valid loadable scene count as zero.}
  \label{fig:supp-checkpoints}
  \vspace{-15pt}
\end{figure}

\paragraph{Efficiency and time-budget trajectories.}
Figure~\ref{fig:supp-efficiency} shows that Claude Opus 5.5 completes tasks fastest at 90.7\,min with 121.7k output tokens while GPT-6 Astra writes the most concise build scripts at 58.9k output tokens and 326 code lines, whereas DeepSeek-V4.1-Flash-1017 and GLM 5.3 Flash emit 304.0k to 329.8k output tokens over 109.0 to 153.5\,min. In Figure~\ref{fig:supp-checkpoints}, where unbuilt or unloadable early snapshots count as zero, GPT-6 Astra produces valid scenes rapidly at 23.1 by 15\,min before plateauing near 39.4, while Claude Opus 5.5 climbs from 22.7 at 30\,min to 58.3 through sustained closed-loop refinement.

\subsection{Real-World Robot Validation Details}

\label{sec:supp-real-world}

\begin{table}[t]
  \centering
  \caption{\textbf{Held-out simulation success rates across $\pi_{0.5}$ checkpoints.} Percentage success over 100 held-out seeds per task within each policy's own training scene, with deployed 10{,}000-step checkpoints shaded.}
  \label{tab:supp-sim2real-sim-eval}
  \scriptsize
  \riftColSep{0.75pt}
  \renewcommand{\arraystretch}{1.18}
  \begin{tabularx}{\columnwidth}{@{}l<{\hspace{5pt}}r>{\hsize=1.09\hsize\linewidth=\hsize\centering\arraybackslash}X>{\hsize=0.88\hsize\linewidth=\hsize\centering\arraybackslash}X>{\hsize=1.09\hsize\linewidth=\hsize\centering\arraybackslash}X>{\hsize=0.94\hsize\linewidth=\hsize\centering\arraybackslash}X@{}}
    \toprule
    \raisebox{0.5\normalbaselineskip}{\textbf{Training Scene}} & \raisebox{0.5\normalbaselineskip}{\textbf{Steps}}
    & \shortstack{\textbf{Open}\\\textbf{Drawer}\,\uparr}
    & \shortstack{\textbf{Open}\\\textbf{Door}\,\uparr}
    & \shortstack{\textbf{Close}\\\textbf{Drawer}\,\uparr}
    & \shortstack{\textbf{Mean}\\\%\,\uparr} \\
    \midrule
    \emph{Agent-reconstructed} & 5,000  & 90.0 &  9.0 & 11.0 & 36.7 \\
    \rowcolor{rifttabhi}
    \emph{Agent-reconstructed} & 10,000 & 100.0 & 43.0 & 37.0 & 60.0 \\
    \emph{Agent-reconstructed} & 15,000 & 100.0 & 86.0 & 27.0 & 71.0 \\
    \midrule
    \emph{Human-refined} & 5,000  & 52.0 & 11.0 & 11.0 & 24.7 \\
    \rowcolor{rifttabhi}
    \emph{Human-refined} & 10,000 & 82.0 & 38.0 & 31.0 & 50.3 \\
    \emph{Human-refined} & 15,000 & 86.0 & 64.0 &  3.0 & 51.0 \\
    \bottomrule
  \end{tabularx}
  \vspace{-15pt}
\end{table}

\paragraph{Scale priors, hardware, and paired policy training.}
In practical robot deployment, coarse outer tabletop and cabinet dimensions are trivially obtainable within seconds from a tape measure or product specification without depth sensors, 3D scanning, or camera calibration. Prompts supply only these lightweight outer bounds of $37\times34\times53\,\text{cm}$ in Scene~A and $13\times15\times11.5\,\text{cm}$ atop $24\times16.5\times14.5\,\text{cm}$ in Scene~B to anchor a safe metric workspace on the fixed-base platform, leaving 3D placement, part splits, joint axes, interior cavities, and handle geometry entirely to single-image agent reconstruction before attaching calibrated chest and dual-wrist RealSense D435i cameras on the bimanual Franka Research~3 after scene freeze, demonstrating that a single uncalibrated photograph paired with effortless dimension priors effectively boosts zero-shot real-robot manipulation. Under paired random perturbations of $\pm 1\,\text{cm}$ position, $\pm 3^\circ$ yaw, and $5\text{--}7\,\text{cm}$ initial drawer opening, a contact-dynamics IK planner collects 300 trajectories totaling 54{,}058 frames per variant with $\le 3\,\text{mm}$ penetration to fine-tune $\pi_{0.5}$~\cite{intelligence2025pi05} for 15{,}000 steps under four-GPU FSDP at batch size 16 and cosine learning rate $10^{-5}\to 10^{-6}$.

\paragraph{Checkpoint selection and hardware transfer.}
Table~\ref{tab:supp-sim2real-sim-eval} evaluates 100 held-out seeds per task within each policy's own training scene across 1{,}800 episodes to select converged checkpoints for hardware deployment, requiring target contact and $1\,\text{s}$ sustained opening of $\ge 8\,\text{cm}$ for drawers, $\ge 0.70\,\text{rad}$ for doors, or $\le 5\,\text{mm}$ residual gap for closing without clipping out-of-range joint commands. Equally converged 10{,}000-step checkpoints transfer to unified hardware trials in Table~\ref{tab:sim2real} of the main paper, where unrefined geometric and axis errors surface directly as handle-grasp offsets and premature limit stops.

\subsection{Human Study Protocol and Agreement Statistics}
\label{sec:supp-human-study}

\paragraph{Blind pairwise protocol.}
We conduct a double-blind pairwise human study across 10 MuJoCo benchmark tasks and all six native coding agents, forming 15 model pairs per photograph and 150 distinct pairs across 60 submissions. Five raters each evaluate four tasks totaling 60 pairs via an offline browser interface so every pair receives two independent judgments, yielding 300 pairwise comparisons and 600 question-level votes retained without post-hoc adjudication. Each submission is rendered as an anonymized six-row evidence sheet covering camera overlays, collision geometry, closed-and-open top-down layouts, per-part articulation, an oblique overview, and simulated Franka Panda rollouts, with randomized left-right order per rater. Raters independently judge \emph{Scene Reconstruction Quality} in Q1 and \emph{Robot Manipulation Readiness} in Q2 via win, loss, or tie votes, and models are ranked by decisive win rate over non-tied votes.

\paragraph{Alignment with automated RIFT metrics and rater agreement.}
Table~\ref{tab:human-study} in the main paper compares all 300 pairwise votes per question against automated \projectname{} scores evaluated on the identical 60 submissions, where $S_{\text{frozen}}=S_{\text{rebuilt}}$ across all six agents. On Q1 scene quality, automated \projectname{} scores descend monotonically from 66.22 on Claude Opus 5.5 to 12.81 on MiniMax M3.1-Flash in exact six-of-six lockstep with human decisive win rates from 92.59\% to 5.33\%, achieving Spearman $\rho=1.00$ and Kendall $\tau=1.00$ at $p=0.0028$, with Pearson $r=0.93$ against automated pairwise win rates. On Q2 robot readiness, \projectname{} scores maintain $\rho=0.94$ at $p=0.0048$, while GPT-6 Astra's lead over Claude Opus 5.5 at 91.23\% versus 71.64\% mirrors higher automated Franka \emph{Robot Open} success of 93.8\% versus 68.8\% across the 16 scored parts in this subset. Meanwhile, two-rater exact agreement drops from 69.3\% across 104 of 150 pairs with $\kappa$ $[0.396,0.686]$ on Q1 to 48.0\% across 72 of 150 pairs with $\kappa$ $[-0.147,0.384]$ and three negative-$\kappa$ rater pairs on Q2, confirming that human raters struggle to diagnose sub-centimeter clearances and grasp physics visually without inspecting raw rationales and tie distributions, and underscoring the necessity of \projectname{}'s automated physical evaluation.

\end{document}